\documentclass[5  p,twocolumn]{elsarticle}
\usepackage{lineno,hyperref}
\usepackage{amsmath}
\modulolinenumbers[5]
\usepackage{amssymb}
\usepackage{graphicx}
\usepackage{atbegshi}
\usepackage{textcomp}
\usepackage{subcaption}
\usepackage{lipsum}
\usepackage{verbatim}
\usepackage{adjustbox}
\usepackage{multirow}

\usepackage{subcaption}
\usepackage{enumitem}
\usepackage{hyperref}
\usepackage{comment}
\usepackage{booktabs}
\usepackage{enumitem}
\usepackage{hyperref}
\usepackage{comment}
\usepackage{booktabs}
\usepackage{graphicx}
\usepackage{url}
\usepackage{amssymb}
\usepackage{pifont}
\usepackage{enumitem}
\usepackage{enumitem}
\usepackage{array}
\usepackage{times}
\usepackage{soul}
\usepackage{url}
\usepackage{graphicx}
\usepackage{amsmath}
\usepackage{amsthm}
\usepackage{booktabs}
\usepackage{xcolor}
\definecolor{mygreen}{RGB}{0,150,0}
\definecolor{myred}{RGB}{255,0,0}

\definecolor{paired-light-blue}{RGB}{198, 219, 239}
\definecolor{paired-dark-blue}{RGB}{49, 130, 188}
\definecolor{paired-light-orange}{RGB}{251, 208, 162}
\definecolor{paired-dark-orange}{RGB}{230, 85, 12}
\definecolor{paired-light-green}{RGB}{199, 233, 193}
\definecolor{paired-dark-green}{RGB}{49, 163, 83}
\definecolor{paired-light-purple}{RGB}{218, 218, 235}
\definecolor{paired-dark-purple}{RGB}{117, 107, 176}
\definecolor{paired-light-gray}{RGB}{217, 217, 217}
\definecolor{paired-dark-gray}{RGB}{99, 99, 99}
\definecolor{paired-light-pink}{RGB}{222, 158, 214}
\definecolor{paired-dark-pink}{RGB}{123, 65, 115}
\definecolor{paired-light-red}{RGB}{231, 150, 156}
\definecolor{paired-dark-red}{RGB}{131, 60, 56}
\definecolor{paired-light-yellow}{RGB}{231, 204, 149}
\definecolor{paired-dark-yellow}{RGB}{141, 109, 49}

\definecolor{bg1}{HTML}{FF9966}
\definecolor{bg2}{HTML}{CCE5FF}
\definecolor{bg3}{HTML}{FFCC99}
\definecolor{bg4}{HTML}{FFC107}
\definecolor{bg5}{HTML}{FFCCCC}
\definecolor{bg6}{HTML}{D5E8D4}
\definecolor{bg7}{HTML}{eeeeee}
\definecolor{bg8}{HTML}{cdeb8b}
\definecolor{bg9}{HTML}{dae8fc}
\definecolor{bg10}{HTML}{a2e6eb}

\definecolor{bg31}{HTML}{FFCDD2} 

\definecolor{bg32}{HTML}{F8BBD0}

\definecolor{bg33}{HTML}{E1BEE7} 

\definecolor{bg34}{HTML}{D7CCC8} 

\definecolor{bg35}{HTML}{B2DFDB} 

\definecolor{bg36}{HTML}{A5D6A7} 

\definecolor{bg37}{HTML}{FFF9C4} 

\definecolor{bg38}{HTML}{FFECB3} 

\definecolor{bg111}{HTML}{CB6843}

\definecolor{bg112}{HTML}{D77C5C}

\definecolor{bg113}{HTML}{E28E6E}
\definecolor{bg114}{HTML}{E89F7D}
\definecolor{bg115}{HTML}{EDAE8A}
\definecolor{bg116}{HTML}{F0BA95}
\definecolor{bg117}{HTML}{F3C29F}
\definecolor{bg118}{HTML}{F6CCAA}
\definecolor{bg119}{HTML}{F8D5B3}
\definecolor{bg120}{HTML}{FADCBD}
\definecolor{bg121}{HTML}{FCE6C7}

\definecolor{bg39}{HTML}{FFE0B2} 

\definecolor{bg40}{HTML}{3CB371} 

\definecolor{bg43}{HTML}{ffe5d9}

\definecolor{bg15}{HTML}{7FFFD4}

\definecolor{bg17}{HTML}{F0FFFF}

\definecolor{bg18}{HTML}{F5FFFA}

\definecolor{bg19}{HTML}{F8F8FF}

\definecolor{bg20}{HTML}{FFFFFF}

\definecolor{bg21}{HTML}{E1F5FE}

\definecolor{bg22}{HTML}{B3E5FC}

\definecolor{bg23}{HTML}{81D4FA}

\definecolor{bg24}{HTML}{4FC3F7}

\definecolor{bg25}{HTML}{29B6F6}

\definecolor{bg26}{HTML}{03A9F4}

\definecolor{bg27}{HTML}{039BE5}

\definecolor{bg28}{HTML}{0288D1}

\definecolor{bg29}{HTML}{0277BD}

\definecolor{bg30}{HTML}{01579B}

\definecolor{bg16}{HTML}{FFCC99}

\definecolor{pg51}{HTML}{E8F5E9} 
\definecolor{pg52}{HTML}{C8E6C9} 
\definecolor{pg53}{HTML}{B9F6CA} 
\definecolor{pg54}{HTML}{A9DFBF} 
\definecolor{pg55}{HTML}{BCF5A6} 

\definecolor{pg56}{HTML}{BEF1CE} 
\definecolor{pg57}{HTML}{CEF6EC} 
\definecolor{pg58}{HTML}{B7F0B1} 
\definecolor{pg59}{HTML}{B1F2B5} 
\definecolor{pg60}{HTML}{9DF3C4} 

\definecolor{pg61}{HTML}{DEF7E0} 
\definecolor{pg62}{HTML}{E8F8DC} 

\definecolor{pg63}{HTML}{EBF7E7} 
\definecolor{pg64}{HTML}{F0FDF4} 

\definecolor{pg65}{HTML}{F1FEE7} 
\definecolor{pg66}{HTML}{F7FFF6} 
\definecolor{pg67}{HTML}{FCFFE7} 
\definecolor{pg68}{HTML}{F4FFD2} 
\definecolor{pg69}{HTML}{EEFFE2} 
\definecolor{pg70}{HTML}{E3FDF5} 

\definecolor{connect-color}{RGB}{0,0,0}
\definecolor{middle-color}{RGB}{255,255,255}
\definecolor{leaf-color}{RGB}{173,216,230}
\definecolor{line-color}{RGB}{25,25,112}

\usepackage{algorithm}
\usepackage{algorithmic}

\usepackage{times}
\usepackage{soul}
\usepackage{url}
\usepackage[utf8]{inputenc}
\usepackage{caption}
\usepackage{graphicx}
\usepackage{amsmath}
\usepackage{amsthm}
\usepackage{booktabs}
\usepackage{algorithm}
\usepackage{algorithmic}

\usepackage{url}            
\usepackage{booktabs}       
\usepackage{amsfonts}       
\usepackage{nicefrac}       
\usepackage{microtype}      
\usepackage{xcolor}         
\usepackage{amsmath}
\usepackage{graphicx}

\definecolor{hidden-draw}{RGB}{20,68,106}
\definecolor{hidden-pink}{RGB}{255,245,247}
\definecolor{red}{RGB}{255,0,0}

\definecolor{hidden-draw}{RGB}{0,0,0}
\definecolor{hidden-pink}{RGB}{255,182,193}
\usepackage{forest}    
\usepackage{hyperref}  
\usepackage{tikz}      

\tikzset{
    root style/.style={
        draw,
        rounded corners,
        fill=blue!30, 
        align=center,
        font=\bfseries
    },
    child style/.style={
        draw,
        rounded corners,
        fill=green!30, 
        align=center,
        font=\bfseries
    },
    grandchild style/.style={
        draw,
        rounded corners,
        fill=red!30, 
        align=center,
        font=\bfseries
    }
}

\tikzset{
  my-box/.style={
    rectangle,
    draw=hidden-draw,
    rounded corners,
    text opacity=1,
    minimum height=1.5em,
    minimum width=40em,
    inner sep=2pt,
    align=center,
    line width=0.8pt,
  },
  leaf/.style={
    my-box,
    minimum height=1.5em,
    text=black,
    align=center,
    font=\normalsize,
    inner xsep=2pt,
    inner ysep=4pt,
    line width=0.8pt,
  }
}

\journal{Journal of Knowledge Based System}

\AtBeginDocument{%
  \providecommand\BibTeX{{%
    \normalfont B\kern-0.5em{\scshape i\kern-0.25em b}\kern-0.8em\TeX}}}

\begin{document}

\begin{frontmatter}

\title{Exploring the Frontier of Vision-Language Models: A Survey of Current Methodologies and Future Directions}

\author{Akash Ghosh$^{1}$\corref{mycorrespondingauthor}}
\cortext[mycorrespondingauthor]{Corresponding author}
\ead{akash_2321cs19@iitp.ac.in}
\author{Arkadeep Acharya$^{1}$}
\author{Sriparna Saha$^{1}$}
\author{Vinija Jain$^{2,3}$}
\author{Aman Chadha$^{2,3}$}

\address{$^{1}$Department of Computer Science and Engineering, IIT Patna, India\\ 
$^2$Stanford University, $^3$Amazon AI\\}

\begin{abstract}
The emergence of Large Language Models (LLMs) has profoundly altered the course of the AI revolution. Nevertheless, these LLMs exhibit a notable limitation, as they are primarily adept at processing textual information. To address this constraint, researchers have endeavored to integrate visual capabilities with LLMs, resulting in the emergence of Vision-Language Models (VLMs).These sophisticated models play a crucial role in addressing complex tasks like generating captions for images and responding to visual questions. In our comprehensive survey paper, we delve into the key advancements within the realm of VLMs. Our classification organizes VLMs into three distinct categories: models dedicated to vision-language understanding, models that process multimodal inputs to generate unimodal (textual) outputs and models that both accept and produce multimodal inputs and outputs.
This classification is based on their respective capabilities and functionalities in processing and generating various modalities of data.We meticulously dissect each model, offering an extensive analysis of its foundational architecture, training data sources, as well as its strengths and limitations wherever possible, providing readers with a comprehensive understanding of its essential components. We also analyzed the performance of VLMs in various benchmark datasets. By doing so, we aim to offer a nuanced understanding of the diverse landscape of VLMs. Additionally, we underscore potential avenues for future research in this dynamic domain, anticipating further breakthroughs and advancements.

\end{abstract}

\end{frontmatter}

\section{Introduction}
The rise of Large Language Models (LLMs) signifies the dawn of a transformative period in Artificial Intelligence, fundamentally restructuring the entire domain. Research labs, spanning both academia and industry, are actively engaged in a competitive race to advance the capabilities of LLMs. Yet, a notable limitation has come to the forefront – these models are confined to processing a singular modality of data, specifically text. This constraint underscores a pivotal challenge in the ongoing pursuit of refining LLMs to operate seamlessly across multiple modalities, marking a crucial avenue for further innovation in the realm of AI.\par
Natural intelligence excels in processing information across multiple modalities, encompassing written and spoken language, visual interpretation of images, and comprehension of videos. This innate ability to seamlessly integrate diverse sensory inputs enables humans to navigate the complexities of the real world. For artificial intelligence to emulate human-like cognitive functions, it must similarly embrace multimodal data processing. This imperative is not merely technological but essential for providing AI systems with contextual awareness and adaptability in real-world scenarios.\par

In response to these limitations, researchers have pioneered a cutting-edge class of neural models known as Vision-Language Models (VLMs). These models intricately combine visual and textual information, showcasing a remarkable proficiency in comprehending and generating content that involves both images and text. Designed to excel in tasks such as responding to visual queries\cite{liu2023llava,li2023blip,li2023blip2,penamakuri2024visual,penamakuri2025big}, image captioning \cite{bai2023qwenvl,ye2023mplugowl} ,multimodal summarization\cite{ghosh2024clipsyntel,ghosh2024medsumm,ghosh2024sights,ghosh2024healthalignsumm} , generating infographics\cite{ghosh2025infogen} , multimodal retrieval \cite{penamakuri2023answer} etc . VLMs exhibit a versatile capability. Their seamless integration of visual and linguistic modalities positions them at the forefront of technological advancements, allowing them to navigate the intricate interplay between images and text with unparalleled finesse.\par

\begin{figure*}[t]
  \begin{minipage}{0.5\textwidth}
    \centering
    \vspace{2mm}    
    \includegraphics[width=\linewidth]{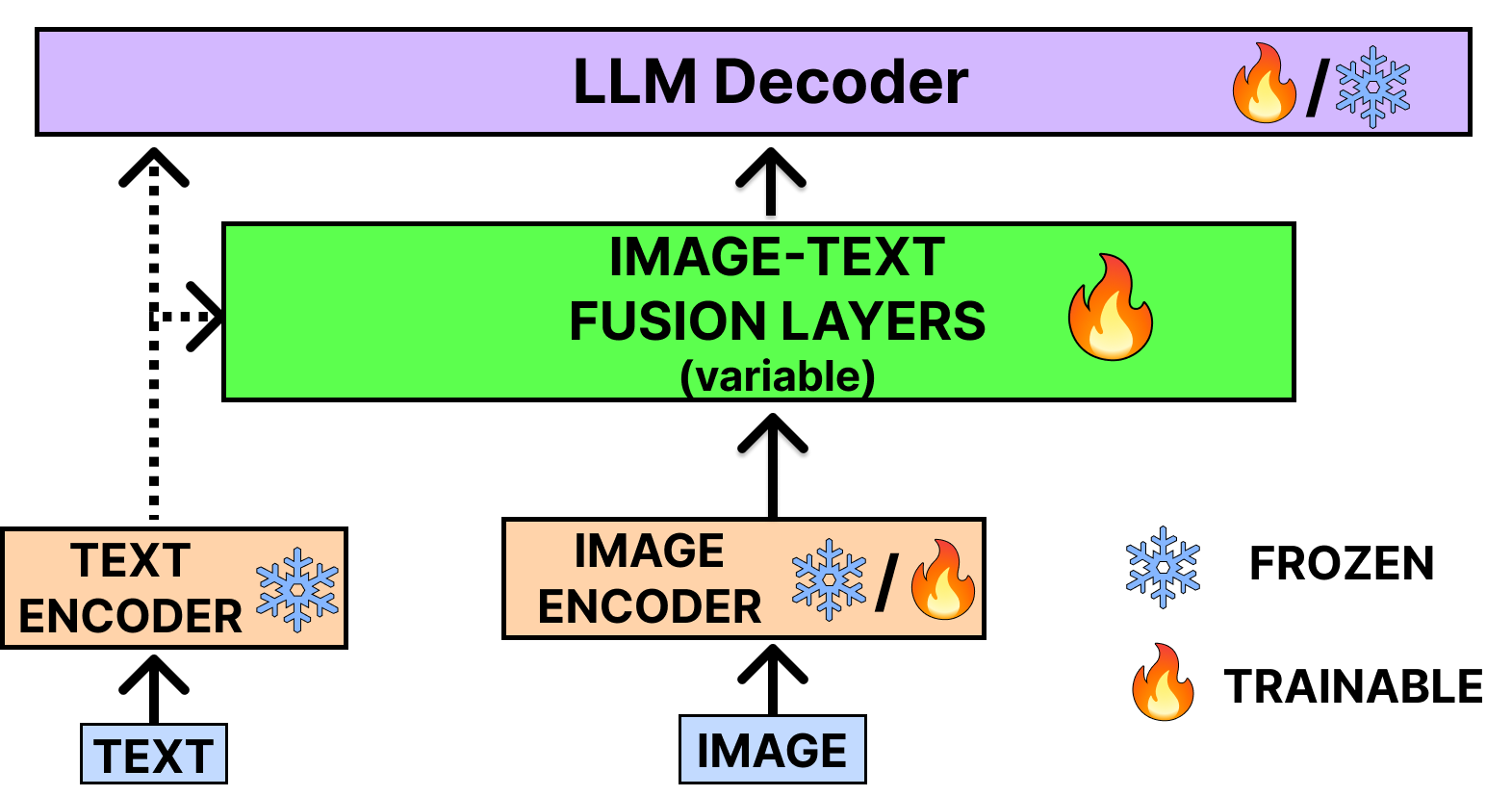}
    \label{fig:minipage}
  \end{minipage}%
  \hspace{0.01\textwidth}
  \begin{minipage}{0.5\textwidth}
  \centering
    \begin{tabular}{|p{50pt}|p{60pt}|p{60pt}|}
        \hline
        \textbf{Component} & \textbf{Trainable} & \textbf{Frozen} \\
        \hline
        \hline
        LLM \newline Decoder & MiniGPT-V2 \cite{chen2023minigptv2}& Fuyu \cite{fuyu-8b}, Qwen-VL \cite{bai2023qwenvl}\\
        \hline
        Image \newline Encoder & Qwen-VL \cite{bai2023qwenvl} & MiniGPT-V2 \cite{chen2023minigptv2}\\
        \hline
    \end{tabular}\\
    \vspace{2mm}
    \textbf{Image-Text Fusion:} \\
    Some prominent examples are:
    \begin{itemize}[itemsep=1pt]
        \item \textbf{Qformer:} BLIP-2 \cite{li2023blip2}
        \item \textbf{Perceiver Resampler:} Flamingo \cite{alayrac2022flamingo}
        \item \textbf{Full Connected Layers (MLP):} Llava \cite{liu2023llava}

    \end{itemize}
    
\end{minipage}

 \caption{A high-level overview of the architecture of VLMs highlighting various design choices with corresponding examples.}
  \label{fig:twocolumnfigure}
\end{figure*}

In recent times, major research labs have been consistently introducing innovative VLMs, including DeepMind's Flamingo, Salesforce's BLIP, and OpenAI's CLIP. Examples like GPT-4 (V) and Gemini showcase how chatbots are evolving within the realm of VLMs. Notably, not all multimodal models are VLMs; for instance, text-to-image models like Midjourney and DALL-E \cite{ramesh2021zero} lack a language generation component, underscoring the diverse landscape of multimodal AI  landscape. The general architecture of a VLM consists of an image and text encoder to generate the embeddings which are then fused in an image-text fusion layer and this fused vector is passed through an LLM to generate the final visually aware generated text. The working of a VLM is shown in  Figure-\ref{fig:twocolumnfigure} \par
In this survey paper, we have categorized VLMs based on their input processing and output generation capabilities into three distinct groups: Vision-Language Understanding Models, Multimodal Input Text Generation models, and the most advanced Multimodal Input-Multimodal Output Models. The subsequent sections delve into comprehensive explanations of each category, elucidating the nuanced functionalities and capabilities of these diverse VLM frameworks.\par
Recent surveys in this area like \cite{wang2023large} mostly explore various pre-trained techniques and datasets used for developing multimodal models, \cite{yin2023survey} explores various key techniques in training various multimodal language models. \cite{wu2023multimodal} provides practical applications and guidance in using multimodal language models. The most recent one by \cite{zhang2024mm} covers around 26 most recent VLMs in depth. In contrast to previous surveys, none have systematically classified Vision-Language Models (VLMs) based on their input-processing and output-generation capabilities. Our survey addresses this gap by providing a thorough categorization of VLMs, shedding light on the intricacies of their functionalities. We extensively analyze the performance of different VLMs across benchmark datasets, notably including the latest MME benchmark, providing comprehensive insights. Our survey represents the most comprehensive and current compilation of VLMs to date, encompassing approximately 70 models. It serves as the ultimate guide for users navigating the evolving realm of vision-language models, offering the most current and comprehensive insights in this pioneering field of research.

\section{Vision-Language Model (VLMs)}

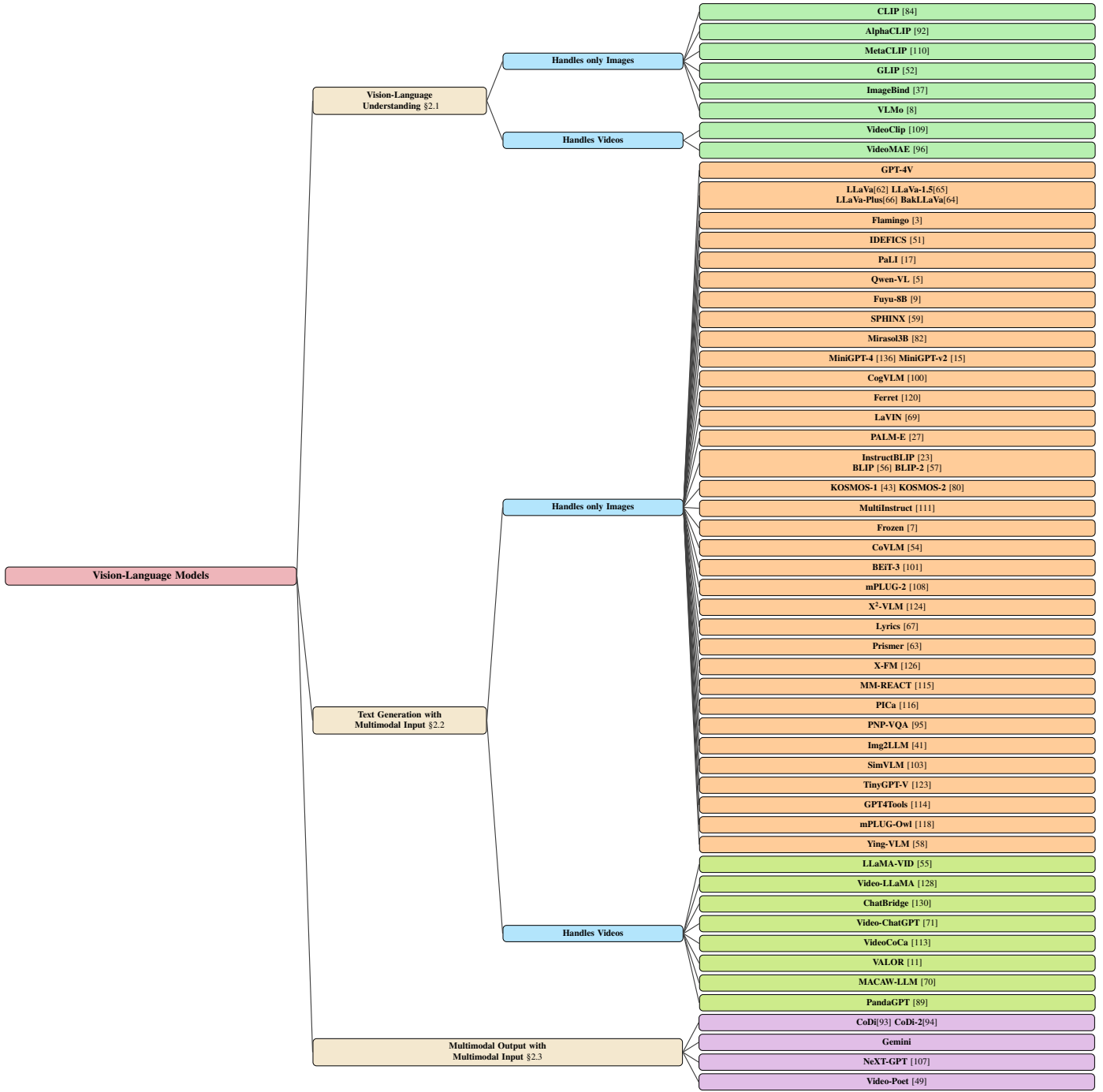
\begin{figure*}[]
  \centering
  \resizebox{\textwidth}{!}{%
  \Large
    \begin{forest}
      for tree={
        grow=east,
        reversed=true,
        anchor=base west,
        parent anchor=east,
        child anchor=west,
        base=center,
        font=\large,
        rectangle,
        draw=hidden-draw,
        rounded corners,
        align=center,
        text centered,
        minimum width=5em,
        edge+={darkgray, line width=1pt},
        s sep=3pt,
        inner xsep=2pt,
        inner ysep=3pt,
        line width=0.8pt,
        ver/.style={rotate=90, child anchor=north, parent anchor=south, anchor=center},
      },
      where level=1{text width=15em,font=\normalsize,}{},
      where level=2{text width=14em,font=\normalsize,}{},
      where level=3{minimum width=10em,font=\normalsize,}{},
      where level=4{text width=26em,font=\normalsize,}{},
      where level=5{text width=20em,font=\normalsize,}{},
      [
        \textbf{Vision-Language Models}, for tree={fill=paired-light-red!70}, text width=22em
        [
          \textbf{Vision-Language}\\ \textbf{ Understanding}   \S\ref{subsec:VLU}, for tree={fill=paired-light-yellow!45}, text width=13em
          [ \textbf{Handles only Images}, for tree={fill=bg22}, text width=13.5em 
          [
            \textbf{CLIP}  \cite{radford2021learning}, for tree={fill=pg58},text width=30em
          ]
          [
            \textbf{AlphaCLIP}   \cite{sun2023alphaclip}, for tree={fill=pg58},text width=30em
          ]
          [
            \textbf{MetaCLIP}   \cite{xu2023demystifying}, for tree={fill=pg58},text width=30em
          ]
          [
            \textbf{GLIP}   \cite{li2022grounded}, for tree={fill=pg58}, text width=30em
          ]
          [
            \textbf{ImageBind}   \cite{girdhar2023imagebind}, for tree={fill=pg58}, text width=30em
          ]
          [
            \textbf{VLMo} \cite{bao2022vlmo}, for tree={fill=pg58},text width=30em
        ]
          ]
          [ \textbf{Handles Videos}, for tree={fill=bg22}, text width=13.5em 
          [
            \textbf{VideoClip}   \cite{xu2021videoclip}, for tree={fill=pg58}, text width=30em
          ]
          [
            \textbf{VideoMAE}   \cite{tong2022videomae}, for tree={fill=pg58}, text width=30em
          ]
        ]
        ]
        [
          \textbf{Text Generation with}\\ \textbf{Multimodal Input}  \S \ref{subsec:TextGenMMInput}, for tree={fill=paired-light-yellow!45}, text width=13em
          [ \textbf{Handles only Images}, for tree={fill=bg22}, text width=13.5em             
          [
                \textbf{GPT-4V} , for tree={fill=bg16},text width=30em
              ]
              [
                \textbf{LLaVa}\cite{liu2023visual}
                \textbf{LLaVa-1.5}\cite{liu2023improved}\\
                \textbf{LLaVa-Plus}\cite{liu2023llavaplus} 
                \textbf{BakLLaVa}\cite{liu2023llava}
                ,text width=30em, for tree={fill=bg16}
              ]
              [
                \textbf{Flamingo}   \cite{alayrac2022flamingo},text width=30em, for tree={fill=bg16}
              ]
              [
                \textbf{IDEFICS}   \cite{laurençon2023obelics},text width=30em, for tree={fill=bg16}
              ]
              [
                \textbf{PaLI}   \cite{chen2023pali}, text width=30em, for tree={fill=bg16}
              ]
              [
                \textbf{Qwen-VL}   \cite{bai2023qwenvl}, text width=30em, for tree={fill=bg16} 
              ]   
              [
                \textbf{Fuyu-8B}   \cite{fuyu-8b}, text width=30em, for tree={fill=bg16}
             ]
             [ 
                \textbf{SPHINX}   \cite{lin2023sphinx}, text width=30em, for tree={fill=bg16}
             ]
             [
                \textbf{Mirasol3B}   \cite{piergiovanni2023mirasol3b}, text width=30em, for tree={fill=bg16}
             ]
             [
                \textbf{MiniGPT-4}   \cite{zhu2023minigpt4} \textbf{MiniGPT-v2}   \cite{chen2023minigptv2}, text width=30em, for tree={fill=bg16}
             ]
             [
                \textbf{CogVLM}   \cite{wang2023cogvlm}, text width=30em,for tree={fill=bg16}
             ]
             [
                \textbf{Ferret}   \cite{you2023ferret}, text width=30em, for tree={fill=bg16}
             ]
             [
                \textbf{LaVIN}   \cite{luo2023cheap}, text width=30em, for tree={fill=bg16}
            ]
            [
                \textbf{PALM-E}   \cite{driess2023palme}, text width=30em, for tree={fill=bg16}
            ]
            [
                \textbf{InstructBLIP}   \cite{dai2023instructblip}\\
                \textbf{BLIP}   \cite{li2022blip} 
                \textbf{BLIP-2}   \cite{li2023blip2}, for tree={fill=bg16}, text width=30em
            ]
            [
                \textbf{KOSMOS-1}   \cite{huang2023language} 
                \textbf{KOSMOS-2}   \cite{peng2023kosmos2}, for tree={fill=bg16},text width=30em
            ]
            [
                \textbf{MultiInstruct}    \cite{xu2022multiinstruct},for tree={fill=bg16},text width=30em
            ]
            [
                \textbf{Frozen}   \cite{bain2021frozen},for tree={fill=bg16},text width=30em
            ]
            [
                \textbf{CoVLM}   \cite{li2023covlm},for tree={fill=bg16},text width=30em
            ]
            [
                \textbf{BEiT-3}   \cite{wang2022image},for tree={fill=bg16}, text width=30em
            ]
            [
               \textbf{mPLUG-2}   \cite{xu2023mplug},for tree={fill=bg16},text width=30em
            ]
            [
               \textbf{X\textsuperscript{2}-VLM}   \cite{zeng2023x2vlm},for tree={fill=bg16},text width=30em
            ]
            [
                \textbf{Lyrics}   \cite{Lu2023LyricsBF},for tree={fill=bg16},text width=30em
            ]
            [
                \textbf{Prismer}   \cite{liu2023prismer},for tree={fill=bg16},text width=30em
            ]
            [
                \textbf{X-FM}   \cite{zhang2023toward},for tree={fill=bg16},text width=30em
            ]
            [
                \textbf{MM-REACT}   \cite{yang2023mm},for tree={fill=bg16},text width=30em
            ]
            [
                \textbf{PICa}   \cite{yang2022empirical},for tree={fill=bg16},text width=30em
            ]
            [
                \textbf{PNP-VQA}   \cite{tiong2022plug},for tree={fill=bg16},text width=30em
            ]
            [
                \textbf{Img2LLM}   \cite{guo2023images},for tree={fill=bg16},text width=30em
            ]
            [
                \textbf{SimVLM} \cite{wang2021simvlm},for tree={fill=bg16},text width=30em
            ]
            [
                \textbf{TinyGPT-V} \cite{yuan2023tinygptv},for tree={fill=bg16},text width=30em
            ]
            [
                \textbf{GPT4Tools} \cite{yang2023gpt4tools},for tree={fill=bg16},text width=30em
            ]
            [
                \textbf{mPLUG-Owl} \cite{ye2023mplugowl}, for tree={fill=bg16},text width=30em
            ]
            [
                \textbf{Ying-VLM} \cite{li2023m3it}, for tree={fill=bg16},text width=30em
            ]
          ]
          [
            \textbf{Handles Videos}, for tree={fill=bg22}, text width=13.5em
            [\textbf{LLaMA-VID}   \cite{li2023llamavid}, text width=30em, for tree={fill=bg8}]
            [
                \textbf{Video-LLaMA}   \cite{zhang2023videollama}, text width=30em, for tree={fill=bg8}
            ]
            [
                \textbf{ChatBridge}   \cite{zhao2023chatbridge}, text width=30em, for tree={fill=bg8}
            ]
            [
                \textbf{Video-ChatGPT}   \cite{maaz2023videochatgpt}, text width=30em, for tree={fill=bg8}
            ]
            [
                \textbf{VideoCoCa}   \cite{yan2023videococa}, text width=30em, for tree={fill=bg8}
            ]
            [
                \textbf{VALOR}   \cite{chen2023valor},for tree={fill=bg8},text width=30em
            ]
            [
                \textbf{MACAW-LLM} \cite{lyu2023macaw},for tree={fill=bg8},text width=30em
            ]
            [
                \textbf{PandaGPT} \cite{su2023pandagpt},for tree={fill=bg8}, text width=30em
            ]
          ]
        ]
        [
            \textbf{Multimodal Output with}\\  \textbf{Multimodal Input}   \S\ref{subsec:MOMI}, for tree={fill=paired-light-yellow!45}, text width=28em
            [
                \textbf{CoDi}\cite{tang2023anytoany}  
                \textbf{CoDi-2}\cite{tang2023codi2}, text width=30em, for tree={fill=bg33}
            ]
            [
                \textbf{Gemini}, text width=30em, for tree={fill=bg33}
            ]
            [
                \textbf{NeXT-GPT}   \cite{wu2023nextgpt}, text width=30em, for tree={fill=bg33} 
            ]
            [
                \textbf{Video-Poet}   \cite{kondratyuk2023videopoet}, text width=30em, for tree={fill=bg33}
            ]
        ]
      ]
    \end{forest}
  }  
  \caption{Taxonomy of Visual Language Models highlighting the input and output formats that the models are capable of handling.}
  \label{fig:lit_surv}
\end{figure*}

This section delves deeply into VLMs, organizing them into three broad classifications for a comprehensive review:
\begin{itemize}
    \item \textbf{Vision-Language Understanding (VLU):} This category encompasses models specifically designed for the interpretation and comprehension of visual information in conjunction with language.
    \item \textbf{Text Generation with Multimodal Input:} Within this classification, we explore models that excel in generating textual content while leveraging multimodal inputs, thereby incorporating diverse forms of information.
    \item \textbf{Multimodal Output with Multimodal Input:} This category delves into models that exhibit proficiency in generating multimodal outputs by processing multimodal inputs. This involves the synthesis of diverse modalities, such as visual and textual elements, to produce comprehensive and coherent results.We have shown this broad classification in the figure-\ref{fig:lit_surv}.\footnote{We have shown the most well-known VLMs in this taxonomy due to space constraints.}\par
\textbf{Comparative Analysis}
We have conducted an extensive analysis of several Vision-and-Language Models (VLMs) across ten widely recognized benchmark datasets, spanning tasks such as Visual Question Answering (VQA) and image captioning. The results of this analysis are presented in Table \ref{tab:res1}. Additionally, we have evaluated the perception and cognition capabilities of these VLMs using the Multimodal Model Evaluation (MME) benchmark, with findings summarized in Table \ref{tab:res2}. Furthermore, a comparative examination of various VLMs on video question-answering datasets is detailed in Table \ref{tab:res3}.

\end{itemize}

\subsection{Vision-Language Understanding}\label{subsec:VLU}
\textbf{CLIP} \cite{radford2021learning}: \label{subsubsec:CLIP}  CLIP, introduced by OpenAI, is a neural network proficient in grasping visual concepts through natural language guidance. It seamlessly identifies visual categories on diverse benchmarks, mirroring the "zero-shot" capabilities seen in GPT-powered models.By scaling a basic contrastive pre-training task, it achieved competitive zero-shot performance on diverse image classification datasets.
CLIP gives a much more robust performance than the fine-tuned deep learning vision models in general for classification tasks. Though CLIP excels at common object recognition but struggles with abstract tasks, fine-grained classification, generalization, and wording sensitivity.\par

\textbf{AlphaCLIP} \cite{sun2023alpha}: The model is an upgraded version of CLIP, incorporating an alpha channel for attentive region indication, enhancing awareness.Alpha-CLIP focuses on specific areas while maintaining CLIP's recognition accuracy using an extended pipeline. It serves as a vision backbone for various applications, excelling in focused region attention but facing challenges with multiple objects and attention amplitude specification. \par

\textbf{MetaCLIP} 
\cite{xu2023demystifying}: The success of CLIP is attributed not only to its model or pre-training objective but also to its rich dataset. Introducing MetaCLIP, an innovative approach, aimed to overcome data transparency issues by refining raw data through metadata sourced from CLIP's concepts. This enhancement demonstrated superior performance over CLIP on benchmarks utilizing a vast collection of 400M image-text pairs sourced from CommonCrawl.\par
\textbf{GLIP} \cite{li2022grounded}: Inspired by CLIP, GLIP employs contrastive pre-training for language-image representations, emphasizing object-level alignment through phrase grounding. It redefines object detection as a vision-language task, utilizing deep fusion for improved representations. GLIP's scalable pre-training on semantic-rich data enables automatic grounding box generation and robust zero/few-shot transferability to outperform baselines like CLIP on image captioning tasks and compete with a fully supervised dynamic head in downstream object detection tasks.

\textbf{VLMo} \cite{bao2022vlmo}:Employing a modular Transformer architecture, VLMo concurrently acquires proficiency in both a dual encoder and a fusion encoder. It adopts a Mixture-of-Modality-Experts (MOME) Transformer framework, which integrates modality-specific experts alongside a shared self-attention mechanism within each module. This design affords substantial modeling flexibility. Vlmo exhibits noteworthy adaptability, functioning adeptly as a fusion encoder for tasks involving vision-language classification and as a dual encoder for efficient image-text retrieval. Its stage-wise pre-training approach optimally utilizes large-scale datasets encompassing images, texts, and their pairs. Consequently, VLMo achieves state-of-the-art performance across diverse vision-language tasks, including visual question answering (VQA) and image retrieval.

\textbf{ImageBind} \cite{girdhar2023imagebind}: Imagebind learns a shared representation space by aligning embeddings from various modalities with image embeddings through diverse paired data sources. This facilitates zero-shot recognition across modalities, leveraging extensive web-scale image-text data and large-scale vision-language models such as CLIP. With minimal training required for deployment across different tasks and modalities, this approach proves highly adaptable. ImageBind utilizes a wealth of large-scale image-text pairs and naturally paired self-supervised data spanning multiple modalities (audio, depth, thermal, IMU) to achieve robust emergent zero-shot classification and retrieval capabilities. It surpasses specialized models on audio benchmarks and demonstrates versatility in compositional tasks. Additional enhancements include the incorporation of richer alignment data and the adaptation of embeddings tailored to specific tasks.\par
\textbf{VideoCLIP} \cite{xu2021videoclip}: VideoCLIP's objective is to pre-train a unified model capable of comprehending both video and text in a zero-shot manner, without dependence on labels for downstream tasks. Its approach involves employing a contrastive learning framework, integrating hard-retrieved negatives and overlapping positives during the pre-training phase for video-text comprehension. Noteworthy innovations include the incorporation of loosely temporally overlapping positive pairs and the utilization of a retrieval-based sampling technique for negative pairs. By leveraging contrastive loss and integrating overlapped video-text clips, VideoCLIP aims to enhance the association between different modalities.It is evaluated on various end tasks, showcasing state-of-the-art performance on video language datasets  like Youcook2\cite{zhou2018towards}. The approach demonstrates significant advancements in zero-shot video-text understanding, outperforming previous work and even supervised approaches in some cases.\par
\textbf{VideoMAE} \cite{tong2022videomae}: VideoMAE is a self-supervised video pre-training method challenging the need for large-scale datasets. It adapts the masked autoencoder framework with a unique video tube masking strategy, achieving data efficiency on small datasets (3k-4k videos). It employs a Vision Transformer with joint space-time attention, demonstrating superior efficiency and effectiveness compared to traditional methods. VideoMAE outperforms in downstream tasks like action detection and holds potential for improvements through dataset expansion and integration of additional data streams. The paper acknowledges energy consumption concerns during pre-training but emphasizes VideoMAE's practical value in scenarios with limited data availability.\par

\newcolumntype{P}[1]{>{\centering\arraybackslash}p{#1}}
\begin{table*}[!ht]
    \centering
    \tiny
    \begin{tabular}{|p{56pt}|P{30pt}|P{30pt}|P{30pt}|P{30pt}|P{30pt}|P{30pt}|P{30pt}|P{30pt}|P{30pt}|P{30pt}|}
    \hline
       \textbf{Model} & \textbf{Science-QA} & \textbf{VizWiz} & \textbf{Flickr30K} & \textbf{POPE} & \textbf{VQA\textsuperscript{v2}} & \textbf{GQA} & \textbf{LlavaBench} & \textbf{Chart-QA} & \textbf{MM-Vet} & \textbf{ViSiTBench}\\ \hline\hline
        LLaVA  & 90.92 & - & - & 50.37 \textsuperscript{R/A} & - & 41.3 & 81.9
        & - & 23.8\textsuperscript{7B} & 1091\textsuperscript{13B} \\ \hline
        LaVIN 13B  & 90.83 & - & - & - & - & - & - & - & - & - \\ \hline
        LaVIN 7B  & 89.41 & - & - & - & - & - & - & - & - & - \\ \hline
        BLIP-2   & 61\textsuperscript{img} & 19.6 & 71.6 & 85.3\textsuperscript{V13B} & 41 & 41 & 38.1\textsuperscript{w} & - & 22.4 & - \\ \hline
        InstructBLIP\textsuperscript{V13B} & 63.1 & 34.5 & 82.8 & 88.57 \textsuperscript{R/A} & - & 49.5 & 58.2\textsuperscript{w} & - & 25.6 & - \\ \hline
        InstructBLIP\textsuperscript{V7B} & 60.5 & - & - & - & - & 49.2 & 60.9\textsuperscript{w} & - & 26.2 & 964 \\ \hline
        MetaCLIP   & 68.77 & - & - & - & - & - & - & - & - & - \\ \hline
        MiniGPT4   & 58.7 & - & - & 79.67 \textsuperscript{R/A} & - & 30.8 & - & - & 24.4 & 900 \\ \hline
        Qwen-VL  & 67.1\textsuperscript{img} & 35.2 & 81 & - & 78.8 & 59.3 & - & - & - & - \\ \hline
        Qwen-VL-Chat& 68.2\textsuperscript{img} & 38.9 & 85.8 & - & 78.2 & 57.5 & - & - & - & - \\ \hline
        GPT-4V  & 85 & 23 & 15.2(S) & - & 0\textsuperscript{VQAS} & 10\textsuperscript{EM} & - & 78.5 & - & 1349 \\ \hline
        LLaVa 1.5 \textsuperscript{V7B}  & 66.8 & 50 & - & 85.9 & 78.5 & 62 & 63.4\textsuperscript{w} & - & - & - \\ \hline
        LLaVa 1.5\textsuperscript{V13B} & 71.6 & 53.6 & - & 85.9 & 80 & 63.3 & 70.7 & - & - & - \\ \hline
        BakLLaVa-1 & 66.7 & - & - & 86.6 & - & - & - & - & - & - \\ \hline
        LLaMa-VID\textsuperscript{V13B}& 70 & 54.3 & - & 86 & 80 & 65 & - & - & - & - \\ \hline
        LLaMa-VID \textsuperscript{V7B} & 68.3 & 54.2 & - & 86 & 79.3 & 64.3 & - & - & - & - \\ \hline
        Sphinx & - & - & - & - & - & - & - & - & - & - \\ \hline
        Gemini & - & - & - & - & - & - & - & 80.8\textsuperscript{u} & - & - \\ \hline
        IDEFICS-9B & - & 35.5 & - & - & 50.9 & 38.4 & - & - & - & 997 \\ \hline
        IDEFICS-80B & - & 36 & - & - & 60 & 45.2 & - & - & - & - \\ \hline
        Flamingo-9B & - & 28.8 & 61.5 & - & 51.8 & - & - & - & - & - \\ \hline
        TinyGPT & - & 33.4 & - & - & - & 33.6 & - & - & - & - \\ \hline
        Flamingo-80B & - & 31.6 & - & - & 56.3 & - & - & - & - & - \\ \hline
        KOSMOS-1 & - & 29.2 & 65.2 & - & 51 & - & - & - & - & - \\ \hline
        Ferret-7B & - & - & 80.39 & 90.24 \textsuperscript{R/A} & - & - & 86.7 & - & - & - \\ \hline
        Ferret-13B & - & - & 81.13 & - & - & - & 87.5 & - & - & - \\ \hline
        mPLUG-Owl & - & - & - & 53.97 \textsuperscript{R/A} & - & - & - & - & - & 1025 \\ \hline
        KOSMOS-2 & - & - & 80.5 & - & 51.1 & - & 70 & - & - & - \\ \hline
        Fuyu-8B & - & - & - & 74.1 & - & - & - & 42.1 & - & - \\ \hline
        LLaMa-Adapter-v2-7b & - & - & - & - & - & - & - & - & 31.4 & 1066 \\ \hline
        PaLM-E\textsuperscript{12B} & - & - & - & - & 76.2 & - & - & - & - & - \\ \hline
        PaLM-E\textsuperscript{562B} & - & - & - & - & 80 & - & - & - & - & - \\ \hline
        Frozen & - & - & - & - & 29.5 & - & - & - & - & - \\ \hline
        Emu2\textsuperscript{37B} & - & - & - & - & 33.3 & - & - & - & - & - \\ \hline 
        Emu2-Chat & - & 54.9 & - & - & 84.9 & 65.1 & - & - & - & - \\ \hline
        Lyrics\textsuperscript{V13B} & 71.1\textsuperscript{img} & 37.6 & 8534 & - & 81.2 & 62.4 & - & - & - & - \\
        \hline
    \end{tabular}
    \caption{Table showing the comparative analysis of multiple Visual Language Models on 10 benchmark datasets: 
    Science-QA \protect\cite{lu2022learn},
    VizWiz \protect\cite{gurari2018vizwiz},
    Flickr30K \protect\cite{plummer2016flickr30k}, 
    POPE \protect\cite{fan2023pope}, 
    VQA\textsuperscript{v2} \protect\cite{goyal2017making},
    GQA \protect\cite{hudson2019gqa},
    LLaVaBench \protect\cite{liu2023llava},
    Chart-QA \protect\cite{masry2022chartqa},
    MM-Vet \protect\cite{yu2023mmvet},
    ViSiTBench \protect\cite{bitton2023visit}
    . \textsuperscript{V}: Vicuna, , \textsuperscript{img}: only image, \textsuperscript{R/A}: Random and Accuracy have been reported, \textsuperscript{VQAS}: VQA Score, \textsuperscript{EM}: EM Accuracy, \textsuperscript{w}: in-the-wild version, \textsuperscript{u}: Ultra version.\textsuperscript{XB}: The model is of X billion parameters.}
    \label{tab:res1}
\end{table*}

\subsection{Text generation with Multimodal Input}\label{subsec:TextGenMMInput}\par
\textbf{GPT-4V} \cite{GPT4(V)}: GPT-4V  marks a significant advancement, allowing users to instruct GPT-4 to analyze image inputs. OpenAI conducted extensive safety evaluations and preparations for GPT-4V, building on the safety work done for GPT-4. The training process involved predicting the next word in a document, utilizing a vast dataset of text and image data. GPT-4V inherits both text and vision capabilities, presenting novel features at their intersection. The system card outlines OpenAI's preparation, early access period, safety evaluations, red team assessments, and implemented mitigations before broad release.\par

\textbf{LLaVA} \cite{liu2023visual}: LLaVA, an open-source multimodal framework designed to enhance LLMs for understanding both language and images. It utilizes language-only GPT-4 to generate data for instruction-following tasks in a multimodal context. LLaVA integrates a vision encoder from CLIP with the LLM, enabling it to process visual information alongside language. The model undergoes pre-training on image-text pairs and fine-tuning for end-to-end multimodal understanding, resulting in a versatile multimodal chatbot. LLaVA demonstrates impressive multimodal chat abilities and achieves an 85.1\% relative score compared to GPT-4 on a synthetic multimodal instruction-following dataset. Upon fine-tuning on the Science QA dataset, LLaVA and GPT-4 together achieve a new state-of-the-art accuracy of 92.53\%.\par

\textbf{Flamingo} \cite{alayrac2022flamingo}: Flamingo introduces novel architectural features to seamlessly integrate vision-only and language-only models. By incorporating interleaved cross-attention layers with frozen language-only self-attention layers, Flamingo excels in handling sequences of intermixed visual and textual data. It adopts a Perceiver-based architecture to convert input sequence data, like videos, into a fixed set of visual tokens. Leveraging large-scale multimodal web corpora with interleaved text and images, Flamingo demonstrates remarkable few-shot learning capabilities across various benchmarks, surpassing models fine-tuned on significantly more task-specific data. This showcases its adaptability and efficiency in rapidly adapting to diverse image and video understanding tasks with limited examples.The OpenFlamingo project is an ongoing effort dedicated to crafting an open-source rendition of DeepMind's Flamingo models. Across a spectrum of seven vision-language datasets, the performance of OpenFlamingo models consistently ranges between 80\% and 89\% when compared to the original Flamingo models. Med-Flamingo \cite{moor2023med}, a medical-focused multimodal few-shot learner based on OpenFlamingo-9B, achieves up to a 20\% improvement in generative medical Visual Question Answering. It pioneers human evaluation in this context, involving clinicians in an interactive assessment, and enables applications like rationale generation.\par

\textbf{PALM-E} \cite{driess2023palm}: PALM-E emerges as an innovative Embodied Multimodal Language Model, meticulously crafted to navigate real-world scenarios by fusing language comprehension with continuous sensor inputs. This model is the result of a collaborative endeavor between TU Berlin and Google Research, signifying a pivotal advancement in the realm of multimodal AI. By integrating continuous sensor modalities from the real world into language models, this approach enables end-to-end training of multimodal sentences using a pre-trained large language model. It effectively addresses various embodied tasks such as robotic manipulation planning, visual question answering, and captioning. PaLM-E, the most extensive model boasting 562 billion parameters, exhibits cutting-edge performance across embodied reasoning tasks and visual-language domains like OK-VQA. Operating on multimodal sentences, it demonstrates the transfer of knowledge from visual-language domains to embodied reasoning tasks, underscoring its adaptability and scalability. PaLM-E faces limitations in relying on low-level language-conditioned policies for robotic tasks, prompting the proposal of self-supervised entity-centric labeling to enhance guidance in intricate tasks.\par

\textbf{BLIP} \cite{li2023blip}: BLIP stands out as an innovative Vision-Language Pre-training framework, overcoming challenges associated with noisy training data. Central to BLIP is its Multimodal Mixture of Encoder-Decoder (MED) architecture, which incorporates Image-Text Contrastive (ITC),Language Modeling (LM) and Image-Text Matching (ITM) objectives during pre-training. Captioning and Filtering (CapFilt) enhances data quality, improving downstream task performance.BLIP, implemented in PyTorch and pre-trained on a diverse 14 million image dataset, demonstrated notable improvements in downstream tasks like image-text retrieval and captioning. Leveraging nucleus sampling and effective parameter sharing, BLIP outperformed existing models on standard datasets.\par

\textbf{BLIP-2} \cite{li2023blip}:BLIP-2 introduces an efficient strategy for vision-language pre-training, utilizing frozen image encoders alongside large language models. The Querying Transformer within BLIP-2 achieves top-tier performance in vision-language tasks while employing fewer parameters, effectively addressing challenges in interoperability among different modality embeddings. A novel addition brought by BLIP-2 is the Querying Transformer (Q-Former), which acts as a trainable link between static image encoders and a fixed LLM. Depicted in Figure[3], the Q-Former undergoes a two-stage pre-training process. Initially, it focuses on learning representations that bridge vision and language, enabling it to comprehend visual elements crucial to accompanying text. Later, the emphasis shifts to generative learning, connecting the Q-Former's output to the fixed LLM and refining its ability to generate visual representations interpretable by the LLM.

\textbf{InstructBLIP} \cite{dai2305instructblip}: InstructBLIP employs instruction-aware visual feature extraction, enhancing its ability to extract informative features tailored to provided instructions. Achieving state-of-the-art zero-shot performance across 13 held-out datasets, InstructBLIP outperforms BLIP-2 and larger models like Flamingo. The models also excel in downstream tasks, demonstrating a 90.7\% accuracy on ScienceQA IMG \cite{lu2022learn}, and showcase qualitative advantages over concurrent multimodal models in diverse capabilities such as  knowledge-grounded image description, visual scene understanding, and multi-turn visual conversation.\par

\textbf{KOSMOS-1} \cite{huang2023language}: KOSMOS-1 is a VLM  from Microsoft. Trained on a web-scale multimodal corpus, KOSMOS-1 excels in language understanding and generation, OCR-free NLP, and various perception-language tasks, showcasing its capabilities in image captioning and visual question answering. Using a Transformer-based architecture, KOSMOS-1 aligns vision with large language models. Its training involves diverse multimodal corpora, including The Pile and Common Crawl, with language-only instruction tuning. Additionally, the model excels in chain-of-thought prompting, generating rationales before addressing complex question-answering tasks. Overall, KOSMOS-1 represents a significant advancement in the field of Multimodal Large Language Models, offering robust performance across a wide range of tasks.\par
\textbf{KOSMOS-2} \cite{peng2023kosmos}: KOSMOS-2 again from Microsoft Research,  advances traditional models by introducing capabilities for perceiving object descriptions, such as bounding boxes, and grounding text in the visual world. Utilizing a unique representation format for referring expressions, KOSMOS-2 links text spans to spatial locations in images. Employing a sophisticated image processing approach, the model combines vision encodings with location tokens to understand and relate specific image areas to textual descriptions. Constructed upon the foundational KOSMOS-1 architecture, this causal language model, based on Transformers, signifies a significant advancement towards Embodiment AI. It marks a pivotal stride towards the integration of language, visual perception, action, and world modeling, bringing them closer together in convergence for  advancing artificial general intelligence. \par
\textbf{MultiInstruct} \cite{xu2022multiinstruct}: MultiInstruct presents a benchmark dataset for multimodal instruction tuning, featuring 62 tasks across 10 categories. Utilizing the OFA pre-trained multimodal language model, the study focuses on enhancing zero-shot performance on diverse tasks through large-scale text-only instruction datasets like Natural Instructions. Results show strong zero-shot performance and reduced model sensitivity to instruction variations. Comparative analysis of transfer learning strategies indicates improved robustness across multimodal tasks. Increasing task clusters during training enhances overall performance, supporting the effectiveness of MultiInstruct.

\textbf{IDEFICS} \cite{laurenccon2023obelisc} : IDEFICS, an open-access reproduction of the closed-source vision-language model Flamingo by DeepMind, boasts 80 billion parameters and is available on HuggingFace. It performs well in image-text benchmarks, such as visual question answering and image captioning, utilizing in-context few-shot learning. IDEFICS has two versions – an 80 billion parameters model and a 9 billion parameters model.
\par
\textbf{PaLI} \cite{chen2022pali}: PALI, or Pathways Language and Image model, from Google Research, leverages large pre-trained encoder-decoder language models and vision transformers for joint language and vision modeling. The model achieves state-of-the-art results in various vision and language tasks across 100+ languages by utilizing a diverse multilingual dataset containing 10B images and texts. With a simple, modular, and scalable design, PaLI highlights the importance of joint scaling in vision and language components for effective training and performance.\par
\textbf{Frozen} \cite{tsimpoukelli2021multimodal}: Frozen, an innovative multimodal learning approach crafted by DeepMind, combines vision encoders trained on image captioning data with a frozen language model. This design empowers the model to swiftly adapt to new tasks in a few-shot setup, showcasing its efficacy across a spectrum of challenges such as visual question answering across diverse benchmark datasets. This approach trains the vision encoder by backpropagating the gradients thorough the frozen language model’s self-attention layers.The system's notable limitation lies in its suboptimal performance on tasks learned with few shots compared to state-of-the-art models using full training sets, highlighting the potential for enhanced zero-shot and few-shot generalization through further improvements in accuracy and reduced seed requirements.

\textbf{Qwen-VL} \cite{bai2023qwen} :  Qwen-VL series, introduced as large vision-language models, encompasses Qwen-VL and Qwen-VL-Chat, demonstrating exceptional performance in tasks such as image captioning, question answering, visual localization, and versatile interactions. Qwen-VLs demonstrate outstanding performance across various vision-centric tasks, surpassing counterparts of similar scale. Their exceptional accuracy extends beyond traditional benchmarks like captioning and question-answering to include recent dialogue benchmarks. Trained on multilingual image-text data, with a significant portion in English and Chinese, Qwen-VLs naturally support multiple languages. They handle multiple images concurrently during training, enabling Qwen-Chat-VL to contextualize and analyze complex scenarios. With higher-resolution inputs and fine-grained training data, Qwen-VLs excel in fine-grained visual understanding, outperforming existing vision-language models in grounding, text comprehension, question answering, and dialogue tasks. \par
\textbf{Fuyu-8B} \cite{fuyu-8b}: Fuyu-8B, a multi-modal text and image transformer developed by Adept AI, offers a simplified yet powerful solution tailored for digital agents. Its straightforward architecture and training process enhance comprehension, scalability, and deployment, making it ideal for various applications. Specifically designed for digital agents, Fuyu-8B seamlessly handles arbitrary image resolutions and excels in tasks like graph and diagram comprehension, UI-based queries, and rapid processing of large images within 100 milliseconds. Despite its optimization for Adept's use case, Fuyu-8B delivers impressive performance in standard image understanding benchmarks such as visual question-answering and natural image captioning. Architecturally, Fuyu adopts a vanilla decoder-only transformer, efficiently processing image patches by linear projection into the first layer. Its versatility in supporting diverse image resolutions is achieved by treating image tokens like text tokens, utilizing raster-scan order, and signaling line breaks for adaptability.\par
\textbf{Sphinx} \cite{lin2023sphinx}:  SPHINX is a versatile VLM that integrates model weights, tuning tasks, and visual embeddings to enhance its capabilities. It unfreezes the large language model during pre-training to strengthen vision-language alignment and efficiently mixes weights from LLMs trained on real-world and synthetic data for robust understanding. By incorporating diverse tasks like region-level understanding and human pose estimation, SPHINX achieves mutual enhancement across different scenarios. It also extracts comprehensive visual embeddings from various sources, enriching language models with robust image representations. SPHINX demonstrates superior multi-modal understanding across applications and introduces an efficient strategy for capturing fine-grained details in high-resolution images, excelling in visual parsing and reasoning tasks.\par
\textbf{Mirasol} \cite{piergiovanni2023mirasol3b}: Mirasol, from Google DeepMind and Google Research, is a multimodal autoregressive model designed to handle both time-aligned modalities (audio, video) and non-aligned modality (text). The architecture involves segmenting long video-audio sequences into manageable chunks, passing them through respective encoders, and using a Combiner to fuse video and audio features. Autoregressive training predicts sequential features, with a separate Transformer block integrating textual prompts through cross-modal attention. This enables enriched contextual understanding, showcasing a comprehensive approach to multimodal learning and generation. Pretrained on 12\% of VTP, the model uniformly weighed losses in pretraining, with a tenfold emphasis on unaligned text loss during fine-tuning. Ablation studies underscore its ability to maintain content consistency and adapt to dynamic changes in video-audio sequence\par

\textbf{MiniGPT-4} \cite{zhu2023minigpt}: MiniGPT-4  combines a frozen visual encoder (ViT + Q-Former from BLIP-2) with   LLM using a single trainable projection layer. Pretrained on aligned image-text pairs and fine-tuned on detailed image descriptions, MiniGPT-4 exhibits GPT-4-like capabilities without training vision or language modules separately. The finetuning process enhances language outputs, demonstrating diverse skills like meme interpretation, recipe generation, and poem composition. The model's architecture involves a vision encoder, linear projection layer, and   large language model.\par
\textbf{MiniGPT-v2} \cite{chen2023minigptv2}: The model architecture of MiniGPT-v2  consists of a ViT visual backbone, a projection layer for dimension matching , and a large language model like LLaMA-2 \cite{touvron2023llama} for the final generation . The ViT backbone is frozen during training, and four adjacent visual output tokens are concatenated and projected into LLaMA-2 space. Task-specific identifiers are incorporated during training using a three-stage strategy with weakly labeled image-text datasets and multi-modal instructional datasets. The model demonstrates superior performance in visual question-answering and visual grounding, outperforming other generalist models. The use of task identifier tokens enhances efficiency in multi-task learning, contributing to its state-of-the-art performance. Challenges include occasional hallucinations, emphasizing the need for more high-quality image-text-aligned data. \par
\textbf{LLaVA-Plus} \cite{liu2023llavaplus}: LLaVA-Plus,  is a general-purpose multimodal assistant designed to enhance LMMs through visual instruction tuning. The model maintains a skill repository with diverse vision and vision-language pre-trained models, activating relevant tools in response to user inputs for various tasks. Trained on multimodal instruction-following data, LLaVA-Plus covers tool usage in visual understanding, generation, and external knowledge retrieval, surpassing its predecessor, LLaVA, in both existing and new capabilities.The training approach involves using GPT-4 to generate instruction data and integrating new tools through instruction tuning, enabling continuous enhancement. LLaVA-Plus demonstrates state-of-the-art performance on VisiT-Bench, a real-life multimodal task benchmark, excelling in tool use compared to other tool-augmented LLMs. \par
\textbf{BakLLaVA} \cite{bitton2023visit}: BakLLaVA represents a Visual Language Model (VLM) crafted through a collaborative endeavor involving LAION, Ontocord, and Skunkworks AI. It harnesses the power of the Mistral 7B base, enhanced by the innovative LLaVA 1.5 architecture. When paired with the llama.cpp framework, BakLLaVA emerges as a swifter and more resource-efficient alternative to GPT-4 with Vision capabilities.\par
\textbf{LLaVa-1.5} \cite{liu2023visual}: It is a refined version of the  LLaVA, focusing on visual instruction tuning to enhance multimodal models.  The paper outlines modifications to LLaVA, such as using CLIP-ViT-L-336px with an MLP projection and incorporating academic-task-oriented Visual Question Answering (VQA) data. Despite its advancements, limitations are acknowledged, such as prolonged training iterations due to the use of full image patches and challenges in processing multiple images and certain domain-specific tasks.\par
\textbf{CogVLM} \cite{wang2023cogvlm}:
CogVLM is an open-source vision-language foundation model developed by researchers from Tsinghua University. Its architecture comprises a Vision Transformer (ViT) encoder (e.g., EVA2-CLIP-E) for image processing, with output mapped into the text feature space using an MLP adapter. The model includes a pre-trained GPT-style language model and a visual expert module added to each layer, consisting of a QKV matrix and an MLP. CogVLM adopts a deep fusion approach, integrating visual and language features at multiple layers through the visual expert module, surpassing traditional shallow alignment methods. Alignment techniques involve pretraining on a vast dataset of 1.5 billion image-text pairs, employing image captioning loss and Referring Expression Comprehension (REC). Fine-tuning on various tasks, with a focus on free-form instructions, leads to the creation of a variant known as CogVLM-Chat.\par
\textbf{FERRET} \cite{you2023ferret}: FERRET is  designed for spatial referring and grounding in images at different shapes and granularities. Ferret's distinct features include a Hybrid Region Representation, blending discrete coordinates and continuous visual features for diverse region inputs. It uses a Spatial-Aware Visual Sampler to handle various region shapes effectively and is trained on the Ground-and-Refer Instruction-Tuning (GRIT) dataset, which includes hierarchical spatial knowledge and hard negative samples. The architecture includes an image encoder, a spatial-aware visual sampler, and a Language Model . Ferret utilizes a pre-trained visual encoder (CLIP-ViT-L/14) and a Language Model's tokenizer for image and text embeddings. Training occurs on the GRIT dataset for three epochs, with the model randomly choosing between center points or bounding boxes to represent regions. In multimodal chatting tasks, Ferret significantly enhances performance by integrating refer-and-ground capabilities.  Notably, Ferret mitigates the issue of object hallucination, a common challenge in multimodal models.\par


\textbf{BARD}\cite{GoogleBARD}: BARD from Google utilizes a reinforcement learning framework to automate machine learning model design, architecture search, and hyperparameter tuning, making it accessible to users without extensive AI expertise. The system is positioned as a standalone experiment, with a focus on productivity, creativity, and curiosity enhancement. Users engage BARD for tasks such as writing resumes, creating workout routines, and planning itineraries. The model is pre-trained on diverse data sources, and responses are generated by considering context, classified against safety parameters, and re-ranked based on quality. Human feedback and evaluation, including fine-tuning and reinforcement learning on human feedback, are used to improve BARD. Limitations include potential inaccuracies, biases, persona attribution, false positives/negatives, and vulnerability to adversarial prompting. Google is committed to addressing these limitations and improving BARD responsibly over time.

\textbf{LLaMA-VID} \cite{li2023llamavid}: LLaMA-VID introduces a novel dual-token strategy, incorporating context and content tokens, to efficiently encode each video frame. This approach enables the model to handle hour-long videos while mitigating computational complexity. LLaMA-VID employs a hybrid architecture, incorporating pre-trained models like  Vicuna for text processing  and a Vision Transformer for image embeddings in videos. The Q-Former introduces the context-attention token (Et) by computing attention between query-generated text embeddings (Q) and visual tokens (X). Et encapsulates relevant visual features. A content token (Ev) is obtained through mean pooling on visual tokens. Both tokens are integrated into the  V decoder for generating text responses. LLaMA-VID's dual-token generation strategy, comprising context and content tokens, ensures adaptability to various settings, optimizing efficiency for videos while preserving detail for single images.LLaMA-VID is a video and image understanding model designed for efficiency, completing training in two days on 8xA100 GPUs. It uses EVA-G for visual encoding and QFormer for text decoding. The training set includes image and video caption pairs, with evaluations on diverse benchmarks. LLaMA-VID excels in zero-shot video QA benchmarks, achieving high accuracy with only two tokens per frame.\par
\textbf{CoVLM} \cite{li2023covlm}:  CoVLM introduces a novel approach to enhance large language models' compositional reasoning by integrating vision-language communicative decoding. Utilizing communication tokens, the model dynamically composes visual entities and relationships, improving language generation through iterative communication with vision encoders and detection networks . CoVLM demonstrates strong performance across a range of tasks including visual reasoning, reading comprehension, and visual question answering. The model represents a noteworthy advancement in integrating vision and language models, with acknowledgment of potential future improvements in compositionality.\par

\textbf{Emu2:} \cite{sun2023generative}: Emu2 stands as a generative multimodal model boasting 37 billion parameters, showcasing exceptional contextual learning across varied multimodal sequences. It sets unprecedented benchmarks in tasks requiring rapid comprehension with limited examples. Employing a unified autoregressive objective, Emu2 seamlessly integrates visual embeddings and textual tokens. Its architecture includes a visual encoder, multimodal modeling, and visual decoder, allowing coherent outputs across different modalities. Emu2 excels in vision-language tasks, instruction tuning, and controllable visual generation, showcasing state-of-the-art performance in image question answering, subject-driven generation, and zero-shot text-to-image generation. The paper acknowledges broader impact considerations and limitations, emphasizing responsible deployment in light of challenges such as hallucination, biases, and question-answering capabilities.\par
\textbf{Video-LLaMA} \cite{zhang2023videollama}: Video-LLaMA is tailored to comprehend both the visual and auditory aspects of videos. By merging pre-trained visual and audio encoders with static LLMs, the model adeptly tackles the complexities of capturing temporal shifts in visual contexts while seamlessly integrating audio-visual cues. Utilizing a Video Q-former for temporal information and an Audio Q-former for audio encoding, the framework aligns audio-visual data with textual information. Experimental results demonstrate Video-LaMA's effectiveness in comprehending video content and generating meaningful responses in audio and video-grounded conversations. However, the paper acknowledges limitations such as restricted perception capacities and challenges with long videos. Despite these, Video-LLaMA represents a notable advancement in audio-visual AI assistants, with the authors providing open-sourced resources for further development.\par
\textbf{Video-ChatGPT} \cite{maaz2023videochatgpt}: It is, a novel multimodal model enhancing video understanding by integrating a video-adapted visual encoder with a Large Language Model. The architecture leverages the CLIP ViT-L/14 visual encoder for spatiotemporal video representations and the  V-v1.1 language model for comprehensive understanding. Notably, a dataset of 100,000 video-instruction pairs is created to fine-tune the model, focusing on temporal relationships and contextual understanding.  The model exhibits competitive performance in correctness, detail orientation, contextual and temporal understanding, and consistency, surpassing contemporary models in zero-shot question-answering tasks. Qualitatively, Video-ChatGPT demonstrates proficiency in various video-based tasks but faces challenges in subtle temporal relationships and small visual details, indicating avenues for future improvement.\par
\textbf{LAVIN} \cite{luo2023cheap}: LAVIN is utilizing Mixture-of-Modality Adaptation (MMA) for cost-effective adaptation of  LLMs to vision-language tasks. LaVIN, with lightweight adapters, achieves competitive performance and superior training efficiency in multimodal tasks like science question answering and dialogue. Remarkably, LaVIN requires only 1.4 training hours and 3.8M trainable parameters. Experimental results on the ScienceQA dataset demonstrate efficiency with comparable performance and reduced training time and storage costs. LAVIN represents a breakthrough in cost-effective adaptation but has limitations, including the potential for incorrect responses and challenges in identifying fine-grained details in images.

\textbf{BEiT-3} \cite{wang2022image}: BEiT-3 represents a pioneering multimodal foundation model, embodying substantial integration across language, vision, and multimodal pretraining domains. Distinguished by its remarkable proficiency in transfer learning across both vision-centric and vision-language tasks, BEiT-3 underscores the advancement of convergence through innovative enhancements in backbone architecture, pretraining methodologies, and scalable model design. Leveraging Multiway Transformers, this model is characterized by a modular architecture facilitating profound fusion capabilities and modality-specific encoding. With a unified backbone, BEiT-3 executes cohesive masked "language" modeling across images, English text, and image-text pairs, colloquially referred to as "parallel sentences". Empirical findings corroborate BEiT-3's attainment of state-of-the-art performance benchmarks across a spectrum of tasks encompassing object detection, semantic segmentation, image classification, visual reasoning, visual question answering, image captioning, and cross-modal retrieval.\par
\textbf{mPLUG-2} \cite{xu2023mplug}: mPLUG-2 leads the way by introducing a multi-module composition network, diverging from the traditional approach of sequence-to-sequence generation. This innovative design fosters modality collaboration while effectively addressing modality entanglement. The flexibility inherent in mPLUG-2 allows for the selective use of diverse modules across text, image, and video modalities for various understanding and generation tasks. Empirical evaluations demonstrate mPLUG-2's prowess, achieving state-of-the-art or competitive results across an extensive range of over 30 downstream tasks. From challenging multi-modal endeavors like image-text and video-text understanding to uni-modal tasks spanning text-only, image-only, and video-only domains, mPLUG-2 exhibits its versatility. A  notable achievement of mPLUG-2 is its groundbreaking performance, achieving a top-1 accuracy of 48.0 and an 80.3 CIDEr score in video QA and caption tasks. Impressively, these results were obtained with a significantly smaller model size and dataset scale. Furthermore, mPLUG-2 exhibits robust zero-shot transferability in both vision-language and video-language tasks, consolidating its forefront status in advancing multimodal pretraining methodologies.\par
\textbf{X\textsuperscript{2}-VLM} \cite{zeng2023x2vlm}: X\textsuperscript{2}-VLM is a versatile model with a flexible modular architecture that integrates image-text and video-text pre-training into a unified framework. It excels in image-text and video-text tasks across different scales, balancing performance and model scale. X\textsuperscript{2}-VLM's modular design enhances transferability, allowing seamless use in various languages or domains. By substituting the text encoder, it outperforms state-of-the-art multilingual multimodal pre-trained models, demonstrating superior performance without requiring specific multilingual pre-training. This adaptability positions X\textsuperscript{2}-VLM as a promising model in the field of multimodal pre-training.\par
\textbf{Lyrics} \cite{Lu2023LyricsBF}:  Lyrics introduces a novel approach to fine-tuning instruction and multimodal pretraining through cross-modal collaboration, built upon the foundational concepts of BLIP-2.It includes advanced techniques for refining visual inputs to extract specific visual characteristics, alongside modules designed for semantic segmentation, object detection, and image tagging. Within the Querying Transformer, visual features seamlessly merge with language inputs, enhanced by boundary boxes and tags derived from the visual refiner.A distinctive aspect of Lyrics is its two-stage training process, which aids in bridging the modality gap by aligning visual-language targets during pretraining. To extract valuable features from tangible objects, it employs a crucial technique called semantic-aware visual feature extraction. The effectiveness of this approach is evidenced by its robust performance across various visual-language benchmark tasks and datasets. \par

\textbf{X-FM} \cite{zhang2023toward}: XFM, a novel general foundation model equipped with one language encoder, one vision encoder, and one fusion encoder, featuring a unique training method. The proposed method incorporates two innovative techniques: halting gradients from vision-language training during language-encoder learning and leveraging vision-language training to guide vision-encoder learning. Extensive experiments on benchmark datasets demonstrate that X-FM outperforms existing general foundation models and performs competitively with or surpasses models tailored specifically for language, vision, or vision-language understanding. The paper acknowledges limitations, including substantial computational requirements, and aims to explore techniques for efficiency improvement and reduced environmental impact. The authors highlight their commitment to addressing efficiency challenges and reducing the carbon footprint in line with "green" deep learning initiatives. However, due to computational constraints, the study did not explore super-large models or pre-train large-sized models on extensive datasets, emphasizing scalability as an essential consideration for foundation models.\par
\textbf{VALOR} \cite{chen2023valor}: VALOR is  a unified vision-audio-language crossmodality pretraining model designed for trimodality understanding and generation. VALOR employs two pretraining tasks, Multimodal Grouping Alignment and Multimodal Grouping Captioning, showcasing good versatility and scalability. Two datasets, namely VALOR-1M and VALOR-32K, emerge as pivotal resources for the advancement of tri-modality pretraining research, aimed at benchmarking audiovisual-language retrieval and captioning. Upon completion of training on the VALOR-1M dataset alongside other vision-language datasets, VALOR establishes novel performance benchmarks across diverse downstream tasks. These tasks encompass retrieval scenarios incorporating vision, audio, and audiovisual inputs, as well as tasks such as captioning and question answering. The documentation delineates prospective avenues for future research, notably including the expansion of the VALOR-1M dataset through unsupervised methodologies, alongside the integration of vision and audio generation modeling within the overarching VALOR framework.\par

\textbf{Prismer} \cite{liu2023prismer}: Prismer is a data- and parameter-efficient vision-language model that utilizes a frozen ensemble of domain experts, minimizing the need for extensive training data. By inheriting weights from pre-trained domain experts across various domains and keeping them frozen during training, Prismer efficiently adapts to different vision-language reasoning tasks. Despite its small-scale language model foundation, Prismer demonstrates competitive fine-tuned and few-shot learning performance, requiring significantly less training data than current state-of-the-art models. However, it lacks the ability for zero-shot in-context generalization and shows limitations in adapting to new experts or partial expert ensembles during inference, leading to performance drops. The paper discusses these limitations, including the absence of few-shot in-context prompting, challenges in adapting to new experts, and potential improvements in representing expert knowledge for enhanced reasoning performance in future iterations.\par

\textbf{MMReact} \cite{yang2023mm}: MM-REACT introduces a novel textual prompt design enabling language models to process multimodal information, including text descriptions, spatial coordinates, and file names for dense visual signals. The approach demonstrates its effectiveness in zero-shot experiments, showcasing its potential for advanced visual understanding across various scenarios. However, the paper identifies limitations, such as challenges in systematically evaluating performance due to the absence of annotated benchmarks for recognition capability in the wild. The integrated vision experts may introduce errors, and the system's success depends on the availability of necessary experts. Additionally, the number of experts is constrained by the context window of ChatGPT, and the conversion of visual signals to text words may not be optimal for certain tasks. Manual prompt engineering is required, and the authors suggest future research to automate this process for increased system development ease.\par
\textbf{PICa} \cite{jin2021good}:  PICa is a method utilizing image captions to prompt GPT-3 for knowledge-based Visual Question Answering (VQA). Leveraging GPT-3's knowledge retrieval and question-answering capabilities, the approach treats GPT-3 as an implicit and unstructured knowledge base, converting images into captions or tags for GPT-3 understanding. By adapting GPT-3 for VQA through a few-shot learning approach with in-context examples, PICa achieves notable performance, surpassing the supervised state of the art on the OK-VQA dataset with just 16 examples. The method is the first to use GPT-3 for multimodal tasks. However, a limitation is noted, as the image is abstracted as text, and captions may provide only a partial description, potentially missing crucial visual details necessary for detailed question answering, such as queries on specific visual attributes.\par

\textbf{PNP-VQA} \cite{tiong2022plug}:  Plug-and-Play VQA (PNP-VQA) is  a modular framework designed for zero-shot Visual Question Answering (VQA). Unlike existing approaches that demand extensive adaptation of pre-trained language models (PLMs) for vision, PNP-VQA eliminates the need for additional training of PLMs. Instead, it employs natural language and network interpretation as an intermediate representation to connect pre-trained models. The framework generates question-guided informative image captions, utilizing them as context for PLMs during question answering.PNP-VQA outperforms end-to-end trained baseline models and establishes new benchmarks by achieving state-of-the-art results on zero-shot VQAv2 and GQA datasets. Despite having 11 billion parameters, it surpasses the performance of an 80 billion-parameter model on VQAv2 and demonstrates a remarkable 9.1\% improvement over a comparable model on GQA. This highlights its effectiveness across a range of parameter sizes for pre-trained language models (PLMs).\par

\textbf{Img2LLM} \cite{guo2023images}: Img2LLM is  designed for LLMs that facilitates zero-shot  VQA without necessitating end-to-end training. The approach involves developing LLM-agnostic models that articulate image content through exemplar question-answer pairs, proving to be effective prompts for LLMs. Img2LLM boasts several advantages, achieving performance on par with or surpassing end-to-end trained methods, such as outperforming Flamingo by 5.6\% on VQAv2 and exhibiting notable superiority on the challenging A-OKVQA dataset. Additionally, the flexibility of Img2LLM allows seamless integration with various LLMs for VQA tasks, eliminating the need for specialized, costly end-to-end fine-tuning. One caveat is the additional inference overhead incurred during image caption and question-answer pair generation, contributing to a 24.4\% increase in computational time. However, this overhead can be mitigated by shortening prompts, trading a fraction of accuracy for speed, while Img2LLM avoids the resource-intensive end-to-end multimodal representation alignment seen in comparable models like Flamingo.\par
\textbf{SimVLM} \cite{wang2021simvlm}: 
SimVLM is a streamlined pretraining framework that embraces a minimalist approach. Unlike previous methods, SimVLM simplifies training complexities by leveraging large-scale weak supervision, and undergoing end-to-end training with a singular prefix language modeling objective. Remarkably, without resorting to additional data or task-specific tailoring, the resultant model surpasses its predecessors like OSCAR, VILLA etc establishing new benchmarks in various vision-language tasks. Additionally, SimVLM demonstrates robust generalization and transfer capabilities, showcasing zero-shot behavior in tasks such as open-ended visual question answering and cross-modality transfer.\par

\textbf{VideoCOCA } \cite{yan2022video}: VideoCoCa is an adaptation of the Contrastive Captioners CoCa \cite{yan2022video} model for video-text tasks. Utilizing CoCa's generative and contrastive attentional pooling layers, VideoCoCa achieves state-of-the-art results in zero-shot video classification and text-to-video retrieval with minimal additional training. The model processes uniformly sampled frames through CoCa's image encoder, creating a tensor representing the entire video sequence. This tensor undergoes attention-pooling layers for both generative and contrastive modeling tasks. VideoCoCa demonstrates proficiency in various video-based tasks, including video reasoning and action recognition, but faces challenges in subtle temporal relationships. Various adaptation strategies and lightweight finetuning approaches were explored, with the attentional pooler method proving most effective. The model was tested on multiple datasets, exhibiting significant improvements over the CoCa baseline. VideoCoCa consistently outperforms CoCa across various scales and tasks, showcasing its robust performance in video-text modeling.

\textbf{TinyGPT-V} \cite{yuan2023tinygpt}: TinyGPT-V addresses the challenges posed by closed-source and computationally demanding multimodal models like GPT-4V. A notable achievement of this model is its impressive performance while utilizing minimal computational resources, requiring only 24GB for training and 8GB for inference. TinyGPT-V, after integrating Phi-2 and vision modes from CLIP, demonstrates competitive performance across various visual question answering and comprehension benchmark datasets when compared to larger models like LLAVA.The model's compact and efficient design, combining a small backbone with large model capabilities, marks a significant step towards practical, high-performance multimodal language models for diverse applications.

\textbf{ChatBridge} \cite{zhao2023chatbridge}: ChatBridge is a multimodal language model aiming to create versatile AI models capable of understanding diverse real-world modalities. The model utilizes language as a conduit, harnessing language-paired data to forge connections between diverse modalities. ChatBridge expands upon the zero-shot capabilities of large language models through a two-stage training procedure, aligning each modality with language and refining with a fresh multimodal instruction dataset . The model demonstrates strong results on zero-shot multimodal tasks, encompassing text, image, video, and audio. Nevertheless, there are constraints in effectively comprehending lengthy videos and audio, indicating a requirement for a more refined temporal modeling method. There is potential to expand the framework by incorporating supplementary modalities such as sketches and point clouds. While employing frozen modules helps mitigate computational constraints, it may also result in inadequate performance and introduce biases inherited from pre-trained models.\par
\textbf{Macaw LLM} \cite{lyu2023macaw}: Macaw-LLM represents a pioneering multi-modal large language model, seamlessly blending visual, audio, and textual data. Its architecture includes a dedicated modality module for encoding multi-modal information, a cognitive module leveraging pre-trained LLMs, and an alignment module harmonizing disparate representations. This alignment facilitates the integration of multi-modal features with textual information, streamlining adaptation processes. Additionally, a comprehensive multi-modal instruction dataset has been curated to support multi-turn dialogue. However, the paper acknowledges certain limitations, particularly regarding the accuracy of the evaluation in fully capturing the capabilities of Macaw-LLM. The model is not optimized for multi-turn dialogues, and potential issues like hallucination, toxicity, and fairness are not evaluated due to the unavailability of suitable evaluation suites.\par
\textbf{GPT4Tools} \cite{yang2023gpt4tools}: GPT4Tools, aiming to enable open-source LLMs, such as LLaMA and OPT, to efficiently use multimodal tools. It addresses challenges posed by proprietary LLMs like ChatGPT and GPT-4, which often rely on inaccessible data and high computational costs. GPT4Tools creates instructional datasets that support large open-source models such as LLAMA in tackling visual challenges through LORA optimization. The approach significantly improves tool invocation accuracy and enables zero-shot capacity for unseen tools. However, the explicit and fixed prompt approach reduces computational efficiency, prompting the exploration of implicit tool invocation methods. Despite limitations, GPT4Tools is considered a viable approach for equipping language models with multimodal tools.\par
\textbf{PandaGPT} \cite{su2023pandagpt}: PandaGPT is an approach enhancing large language models with visual and auditory instruction-following capabilities. PandaGPT excels in tasks like image description, video-inspired story writing, and answering audio-related questions. It seamlessly handles multimodal inputs, connecting visual and auditory information. By combining ImageBind's multimodal encoders and  Vicuna's large language models, PandaGPT requires only aligned image-text pairs for training and exhibits emergent cross-modal behaviors for various data modalities. The paper suggests improvements, including using additional alignment data, exploring fine-grained feature extraction, generating richer multimedia content, creating new benchmarks, and addressing common language model deficiencies. Despite these considerations, PandaGPT represents a promising step toward building Artificial General Intelligence for holistic perception across diverse modalities.\par

\begin{table*}[!ht]
    \tiny
    \centering
    \begin{tabular}{|c|c|c|c|c|c|c|c|c|c|c|c|c|c|c|c|}
    \hline
    \multirow{2}{*}{\textbf{Models}} &
  \multirow{2}{*}{\textbf{Overall}} &
  \multicolumn{10}{c|}{\textbf{Perception}} &
  \multicolumn{4}{c|}{\textbf{Cognition}}\\
   \cline{3-16}  &  &Exist. & Count & Pos. & Color & Poster & Cele. & Scene & Land & Art & OCR & Com. & Cal. & Trans. & Code \\ \hline
    \hline
        Sphinx & 1870.2 & 195 & 160 & 153.3 & 160 & 164.3 & 177.9 & 160 & 168.1 & 134 & 87.5 & 130 & 55 & 75 & 50 \\ \hline
        GPT-4V & 1926.6 & 190 & 160 & 95 & 150 & 192.2 & 0 & 151 & 138.3 & 148 & 185 & 142.1 & 130 & 75 & 170 \\ \hline
        Gemini & 1933.4 & 175 & 131.7 & 90 & 163.3 & 165 & 147.4 & 144.8 & 158.8 & 135.8 & 185 & 129.3 & 77.5 & 145 & 85 \\ \hline
        LLaVa & - & 50 & 50 & 50 & 55 & 50 & 48.82 & 50 & 50 & 49 & 50 & 57.14 & 50 & 57.5 & 50 \\ \hline
        MiniGPT-4 & - & 68.33 & 55 & 43.33 & 75 & 41.84 & 54.41 & 71.75 & 54 & 60.5 & 57.5 & 59.29 & 45 & 0 & 40 \\ \hline
        LaVIN & - & 185 & 88.33 & 63.33 & 75 & 79.79 & 47.35 & 136.75 & 93.5 & 87.25 & 107.5 & 87.14 & 65 & 47.5 & 50 \\ \hline
        InstructBLIP & - & 185 & 143.33 & 66.67 & 153.33 & 123.81 & 101.18 & 153 & 79.75 & 134.25 & 72.5 & 129.29 & 40 & 65 & 57.5 \\ \hline
        BLIP-2 & - & 160 & 135 & 73.33 & 148.33 & 141.84 & 105.59 & 145.25 & 138 & 136.5 & 110 & 110 & 40 & 65 & 75 \\ \hline
        mPLUG-OWL & - & 120 & 50 & 50 & 55 & 136.05 & 100.29 & 135.5 & 159.25 & 96.25 & 65 & 78.57 & 60 & 80 & 57.5 \\ \hline
        Qwen-VL-Chat & 1487.5 & - & - & - & - & - & - & - & - & - & - & - & - & - & - \\ \hline
        LLaVa-1.5\textsuperscript{V7B}& 1510.7 & - & - & - & - & - & - & - & - & - & - & - & - & - & - \\ \hline
        LLaVa-1.5\textsuperscript{V13B}& 1531.3 & - & - & - & - & - & - & - & - & - & - & - & - & - & - \\ \hline
        LLaMA-VID\textsuperscript{V13B}& 1542.3 & - & - & - & - & - & - & - & - & - & - & - & - & - & - \\ \hline
        LLaMA-VID\textsuperscript{V7B}& 1521.4  & - & - & - & - & - & - & - & - & - & - & - & - & - & -\\ \hline
    \end{tabular}
        \caption{Table showing a comparative analysis of various VLMs on the MME Benchmark \protect\cite{fu2023mme}.\textsuperscript{XB}: The model is of X billion parameters. }
    \label{tab:res2}
\end{table*}

\textbf{mPLUG-Owl} \cite{ye2023mplug}: mPLUG-Owl introduces a unique training approach that empowers LLMs with multi-modal capabilities by modularizing the learning process into three key components: a foundational LLM, a visual knowledge module, and a visual abstractor module. Through a two-stage training methodology, this paradigm effectively aligns image and text data, harnessing the support of LLMs while preserving their generation capabilities. Experimental results demonstrate mPLUG-Owl's superior performance in instruction and visual understanding, multi-turn conversation, and knowledge reasoning. The model exhibits unexpected abilities like multi-image correlation and multilingual understanding but has limitations, including challenges in multi-image correlation, limited multilingual training, and mixed performance in OCR of complex scenes. The model also shows potential in vision-only document comprehension, with strengths in tasks like movie review writing and code generation but limitations in other applications, indicating further exploration opportunities in document understanding and downstream applications.\par
\textbf{Ying-VLM} \cite{li2023m3it}: Ying-VLM is trained on  M\textsuperscript{3}IT dataset . Models trained with M\textsuperscript{3}IT show success in following human instructions, providing engaging responses, and achieving strong performance on unseen videos and  tasks in the Chinese language. The analysis indicates that increasing task numbers improves performance, and instruction diversity influences results.M\textsuperscript{3}It consists of 2.4 million instances, including meticulously crafted task instructions spanning across forty different task.\par

\textbf{BLIVA} \cite{hu2023bliva}: BLIVA is a novel multimodal Language Learning Model designed to handle text-rich visual questions, integrating query and patch embeddings. It outperforms existing VLMs like GPT-4 and Flamingo, showing significant improvements in OCR-VQA and Visual Spatial Reasoning benchmarks. BLIVA's architecture includes a Q-Former for instruction-aware visual features and a fully connected projection layer for additional visual information. It demonstrates an overall improvement in the multimodal LLM benchmark (MME) of 17.72\% compared to  InstructBLIP and performs well in real-world scenarios like processing YouTube thumbnail question-answer pairs.\par

\textbf{LLAVA-phi} \cite{zhu2024llava}:  LLaVA-Phi is a highly efficient multi-modal assistant powered by the compact language model, Phi-2. This model demonstrates significant progress in compact multi-modal systems, showing that even smaller models with 2.7B parameters can effectively engage in complex dialogues blending text and visuals, given proper training. LLaVA-Phi excels in various benchmarks covering visual comprehension, reasoning, and knowledge-based perception, suggesting its suitability for real-time interaction scenarios like embodied agents.Significantly, it highlights how smaller language models can attain advanced levels of comprehension and engagement without sacrificing resource efficiency. The training process involves two stages: (1) feature alignment, where a pretrained vision encoder is connected to a language model using a subset of the LAION-CC-SBU dataset, and (2) visual instruction tuning, using a combination of GPT-generated multimodal instruction-following data and VQA data to teach the model to follow multimodal instructions.

\textbf{MoE-LLaVA} \cite{lin2024moe}: MoE-LLaVA is a groundbreaking training strategy for Large Vision-Language Models. Known as MoE-tuning, this innovative approach efficiently manages performance degradation in multi-modal learning and model sparsity by activating only the top k experts during deployment via routers. Despite its architecture comprising 3 billion sparsely activated parameters, MoE-LLaVA achieves comparable or superior performance to state-of-the-art models while minimizing hallucinations in model outputs. Its architecture includes a visual encoder, visual projection layer in the form of MLP, word embedding layer, stacked LLM and MoE blocks. MoE-tuning comprises three stages: MLP training, parameter training excluding the Vision Encoder, and initializing experts in MoE followed by training only MoE layers. Evaluation on various visual understanding datasets demonstrates MoE-LLaVA's efficiency and effectiveness, with extensive ablation studies and visualizations illustrating its effectiveness and offering insights for future research in multi-modal learning systems.\par

\textbf{Yi-VL} \cite{yi-vl}: Yi Vision Language (Yi-VL) is an open-source multimodal model based on the Yi Large Language Model series, excelling in content comprehension, recognition, and multi-round conversations about images. It leads in recent benchmarks, including English and Chinese. Key features include multi-round text-image conversations, bilingual support, strong image comprehension, and fine-grained resolution at 448 $\times$ 448. Yi-VL utilizes the LLaVA architecture, comprising a vision transformer, projection module, and large language model. However, it has limitations such as supporting only visual question answering, accepting a single image input, and potential content generation issues and object identification inaccuracies in complex scenes. Additionally, it operates at a fixed resolution of 448x448, which may result in information loss for low-resolution images and a lack of additional knowledge for higher resolutions.\par

\textbf{Moondream} \cite{MoonDream}:  Moondream is a 1.6 billion parameter model, meticulously crafted by Vikhyatk, emerges from the fusion of SigLIP, Phi-1.5, and the expansive LLaVa training dataset. Representing a significant milestone in AI research, this model is purposefully unveiled for scholarly exploration, underlining its exclusivity for non-commercial endeavors. This amalgamation of cutting-edge techniques and robust data sets underscores a commitment to advancing the frontiers of artificial intelligence, setting a new benchmark for computational prowess and innovation in the field.\par

\textbf{Shikra} \cite{chen2023shikra}: Shikra is a Multimodal Large Language Model designed to bridge the gap in human-like referential abilities within dialogue. Shikra boasts the capability to interpret spatial coordinates through natural language, facilitated by its straightforward architecture consisting of a vision encoder, alignment layer, and LLM. Unlike other models, Shikra doesn't require additional vocabularies or external plugins, allowing for effortless integration of referential dialogue tasks with diverse vision-language tasks. Its performance is notably strong across various tasks such as REC, PointQA, Image Captioning, and VQA, enabling functionalities like offering object coordinates and comparing user-pointed regions. However, it currently supports only English and lacks user-friendliness for non-English speakers. Future work aims to make Shikra multilingual and explore improved coordinate representations for dense object detection and segmentation tasks. Additionally, like most LLMs, Shikra may generate harmful or counterfactual responses.\par
\textbf{BuboGPT} \cite{zhao2023bubogpt}: BuboGPT stands out as a Visual Language Model (VLM) equipped with visual grounding capabilities, aimed at enhancing cross-modal interaction across vision, audio, and language domains. It offers a detailed comprehension of visual elements and other modalities, enabling precise localization of objects in images during response generation. Employing an off-the-shelf visual grounding module based on SAM for entity extraction and mask correspondence in images, alongside a two-stage training approach and an extensive instruction dataset, BuboGPT strives for comprehensive understanding of text-image-audio interactions.Despite facing challenges such as language hallucination and limited capacities in Grounding QA, BuboGPT exhibits remarkable proficiency in understanding multiple modalities and visual grounding tasks, signaling promising advancements in the realm of multi-modal Language Models.\par

\textbf{ChatSpot:} \cite{zhao2023chatspot}: ChatSpot has been introduced as a unified end-to-end multimodal large language model designed to enhance human-AI interaction. It supports diverse interactive forms such as mouse clicks, drag-and-drop, and drawing boxes, providing users with a flexible and seamless interactive experience. The model is built on precise referring instructions, utilizing various reference representations like points and boxes to focus on specific regions of interest. Additionally, a multi-grained vision-language instruction-following dataset is created for training ChatSpot. Experimental results demonstrate its robustness in region referring, showing minimal instances of region referring hallucination even in the presence of box noises. This highlights ChatSpot's capability for precise region referencing and its potential to improve interactive accuracy and efficiency in multimodal large language models.

\textbf{MiniGPT5} \cite{zheng2023minigpt}: MiniGPT-5 introduces an innovative interleaved vision-and-language generation technique, utilizing "generative vokens" to harmonize image-text outputs. Its distinctive two-staged training strategy focuses on description-free multimodal generation, eliminating the need for comprehensive image descriptions. MiniGPT-5 enhances model integrity with classifier-free guidance, resulting in substantial improvements over baseline models like Divter on the MMDialog dataset. It consistently produces superior multimodal results in human assessments conducted on the VIST dataset, showcasing its effectiveness across a range of benchmarks.\par
\textbf{DRESS} \cite{chen2023dress}:DRESS, a sophisticated multimodal language model, leverages Natural Language Feedback (NLF) from Language Models (LLMs) to enhance alignment through interactive engagements, effectively mitigating key limitations seen in current Virtual Language Models (VLMs). NLF is classified into two categories: critique and refinement, aimed at closely aligning with human preferences and bolstering the model's proficiency in multi-turn conversations. Refinement NLF provides constructive suggestions to enhance responses, while critique NLF aids in aligning VLM outputs with human preferences. Through Reinforcement Learning, DRESS is trained to handle the non-differentiable nature of NLF. Empirical findings demonstrate that DRESS facilitates the generation of more beneficial and benign outputs and adeptly learns from feedback during multi-turn interactions, surpassing state-of-the-art VLMs.\par

\textbf{X-InstructBLIP} \cite{panagopoulou2023x}: X-InstructBLIP is a cross-modality framework built upon frozen large language models that integrates various modalities without extensive customization. High-quality instruction tuning data is collected automatically, enabling fine-tuning for different modalities. The model performs comparably to leading-edge counterparts without extensive pre-training or customization. A novel evaluation task, Discriminative Cross-modal Reasoning (DisCRn), has been introduced to assess the model's cross-modal abilities across disparate input modalities. X-InstructBLIP demonstrates emergent cross-modal reasoning despite separate optimization of each modality and outperforms strong captioning baselines across all examined modalities in DisCRn. However, complexities and unanswered questions within each modality highlight challenges and opportunities for future exploration both across and within modalities.\par

\textbf{VILA} \cite{lin2023vila}: VILA, a Visual Language model family, emerges from an enhanced pre-training recipe that systematically augments LLMs towards VLMs. VILA consistently outperforms state-of-the-art models like LLaVA1.5 across main benchmarks, showcasing its superior performance without additional complexities. Notably, VILA's multi-modal pre-training unveils compelling properties such as multi-image reasoning, enhanced in-context learning, and improved world knowledge, marking significant advancements in visual language modeling.

\begin{table*}[!ht]
    \centering
    \begin{tabular}{|c|c|c|c|c|c|c|}
    \hline
        \multirow{2}{*}{\textbf{Models}} &   
  \multicolumn{2}{c|}{\textbf{MSVD-QA}} &
  \multicolumn{2}{c|}{\textbf{MSRTT-QA}} &
  \multicolumn{2}{c|}{\textbf{ActivityNet-QA}}\\
   \cline{2-7}  & Accu. &Score & Accu. & Score & Accu. & Score  \\ \hline
        VideoLLaMa\textsuperscript{V7B} & 51.6 & 2.5 & 29.6 & 1.8 & 12.4 & 1.1 \\ \hline
        LLaMa-Adapter\textsuperscript{L7B} & 54.9 & 3.1 & 43.8 & 2.7 & 34.2 & 2.7 \\ \hline
        VideoChat\textsuperscript{V7B} & 56.3 & 2.8 & 45 & 2.5 & 26.5 & 2.2 \\ \hline
        Video-ChatGPT\textsuperscript{V7B} & 64.9 & 3.3 & 49.3 & 2.8 & 35.2 & 2.7 \\ \hline
        LLaMa-VID\textsuperscript{V7B} & 69.7 & 3.7 & 57.7 & 3.2 & 47.4 & 3.3 \\ \hline
        LLaMa-VID\textsuperscript{V13B} & 70 & 3.7 & 58.9 & 3.3 & 47.5 & 3.3 \\ \hline
    \end{tabular}
        \caption{Comparative analysis of leading models on 4 zero-shot video QA datasets from \protect\cite{li2023llamavid}. Results are reported with two tokens for each frame.}
    \label{tab:res3}
\end{table*}

\subsection{Multimodal Output with Multimodal Input}\label{subsec:MOMI}

    \textbf{The Composable Diffusion (CoDi):} CoDi \cite{tang2023anytoany}  model adopts a multimodal approach using Latent Diffusion Models for text, image, video, and audio. Text processing involves a variational encoder (VAE) with BERT and GPT-2, image tasks use a latent diffusion model (LDM) with a VAE, and audio tasks utilize an LDM with a VAE encoder-decoder for mel-spectrogram representation. CoDi creates a shared multimodal space through cross-modal generation with Joint Multimodal Generation and cross-attention modules. Training involves individual diffusion models with aligned prompt encoders, and CoDi achieves any-to-any generation with a linear number of training objectives.\par

    \textbf{CoDi-2} \cite{tang2023codi2}: CoDi-2 employs a multimodal encoder, ImageBind, with aligned encoders and a multilayer perceptron for modalities projection. It integrates diffusion models (DMs) into the multimodal latent language model (MLLM) for detailed, modality-interleaved generation. The fusion strategy involves projecting multimodal data into a feature sequence, processed by the MLLM, and utilizing DMs for improved generation quality. The alignment method leverages projections from aligned multimodal encoders to enable the MLLM to understand modality-interleaved input sequences, facilitating in-context learning and supporting multi-round interactive conversations. \par

    \textbf{Google Gemini} \cite{gemini}: Gemini models feature a transformative architecture with deep fusion capabilities, excelling in integrating text, image, audio, and video modalities. They surpass GPT-4 in 30 out of 32 benchmarks and are trained on Google's TPU v4 and v5e accelerators for efficient scaling. The multimodal and multilingual training dataset prioritizes quality and safety, with models undergoing Reinforcement Learning from Human Feedback (RLHF). While specific details remain undisclosed, safety evaluations for bias and toxicity are a central part of Gemini's development, involving collaboration with external experts.\par

    \textbf{NExT-GPT} \cite{wu2023next}: NExT-GPT features three stages: Multimodal Encoding, LLM Understanding and Reasoning, and Multimodal Generation. It uses models like ImageBind for encoding and Transformer-based layers for generation. In inference, modal encoders transform inputs, LLM decides on content, and diffusion decoders use signal tokens for synthesis. The system employs Multimodal Alignment Learning to align features and Modality-switching Instruction Tuning (MosIT) for improved LLM capabilities by aligning modal signal tokens with gold captions. The diverse MosIT dataset enhances the model's ability to handle various user interactions effectively.\par
    \textbf{VideoPoet} \cite{wu2023next}:  VideoPoet is a language model designed for high-quality video synthesis with matching audio. The model employs a decoder-only transformer architecture, processing multimodal inputs like images, videos, text, and audio. Utilizing a two-stage training protocol, VideoPoet showcases state-of-the-art capabilities in zero-shot video generation and excels in tasks such as text-to-video and video stylization. Notable features include a Large Language Model backbone, custom spatial super-resolution, and scalability with model size. Human evaluations highlight VideoPoet's superiority in text fidelity, video quality, and motion interestingness. Responsible AI analysis underscores considerations for fairness, emphasizing the model's capabilities in zero-shot editing, task chaining, and maintaining quality across multiple stages in video generation.\par

\section{Future Directions}
\vspace{-0.2cm}
\textbf{Tradeoff between pre-training and modular structure:} A Lot of research is going on to increase the understanding, control, and faithfulness capacities of VLMs by introducing modularity in place of black box pretraining.
\textbf{Incorporating other modalities :}Works are going on for incorporating more finer modalities like gaze/gestures inspired by \cite{cheng2022gaze} which is very important for the educational sector.\par
\textbf{Fine-grained Evaluation of VLMs:} Works are going on for more fine-grained evaluation of VLMs on parameters like bias, fairness etc. DALL-Eval \cite{cho2023dall} and VP-Eval \cite{cho2023visual} are a few works in this direction.\par

\textbf{Causality and Counterfactual Capabilities in VLMs:} A lot of work is done to understand the causal and counterfactual capabilities of LLM, which inspired researchers to explore the same in the realm of VLMs. Cm3  \cite{aghajanyan2022cm3} is one of the earliest works in this domain and there is a lot of buzz in this topic.\par
\textbf{Continual Learning/Unlearning:} A trend is there on efficiently learning continuously without training it from scratch in the VLM space. VQACL \cite{zhang2023vqacl} and Decouple before Interact \cite{qian2023decouple} are some of the earliest works in this domain. Inspired by the concept of knowledge unlearning observed in LLMs \cite{si2023knowledge}, researchers are also delving into similar approaches within the realm of VLMs.\par
\textbf{Efficieny in training:} Efforts are concentrated on developing efficient multimodal models, with notable advancements such as BLIP-2 showing promise. It surpasses Flamingo-80B by 8.7\% in zero-shot VQA-v2, while utilizing significantly fewer trainable parameters (54 times fewer). \par
\textbf{Multilingual grounding of VLMs:} Following the recent surge in multilingual LLMs such as OpenHathi \cite{OpenHathi} and BharatGPT \cite{BharatGPT}, there is a growing momentum towards the development of multilingual vision-language models (VLMs). Palo\cite{maaz2024palo} is the first notable work in this direction. As applications of multilingual reasoning continue to expand \cite{ghosh2025multilingual,gautam2025mind}, Vision–Language Models (VLMs) must be trained to perform complex reasoning across languages.

\par
\textbf{More Domain specific VLMs:} Various domain-specific VLMs, exemplified by projects like MedFlamingo \cite{moor2023med} and SkinGPT \cite{zhou2023skingpt}, have paved the way in their specialized fields. Further endeavors are in progress to craft VLMs specifically tailored for sectors like education and agriculture.\par

\textbf{Safety of VLMs:} Despite rapid gains, today’s VLMs are still brittle: they can be hijacked by adversarial images at inference time and redirected by visual prompt‑injection, including in clinical settings, revealing systemic goal‑hijacking risks \cite{bailey2023image,clusmann2025prompt}.Unified evaluations like MMJ‑Bench show that jailbreak success remains high across models and that defenses often trade off utility, underscoring the need for layered safeguards and task‑aware testing \cite{weng2025mmj}. Compounding this, multimodal hallucination persists across text, image, video, and audio models, calling for fine‑grained diagnostics and standardized mitigation metrics \cite{sahoo2024unveiling}. As a concrete downstream stake, cyberbullying in memes highlights real‑world harms: comparative studies of generative vs. discriminative detectors\cite{jain2023generative} and recent intervention frameworks like \cite{jha2024memeguard,jha2024meme} illustrate both progress and open gaps in robust multimodal moderation. Going forward, we advocate domain‑specific red‑teaming and evaluation protocols (e.g., healthcare, online safety), robustness to cross‑modal prompt injection, and benchmarks that jointly measure safety, faithfulness (anti‑hallucination), and task quality.

\section{Conclusion}

This paper provides a comprehensive survey of the latest developments in the space of VLMs. We categorize VLMs according to their use cases and output-generation capabilities, offering concise insights into the architecture, strengths, and limitations of each model. Additionally, we highlight future directions in the field, informed by recent trends, to provide a roadmap for further exploration in this domain. We trust that this paper will serve as a valuable resource, offering guidance to researchers in the realms of Computer Vision and Natural Language Processing who are actively involved in interesting areas of multimodal learning.

\section{Acknowledgements}
\vspace{-0.2cm}
Akash Ghosh and Sriparna Saha express their heartfelt gratitude to the  SERB (Science and Engineering Research Board ) POWER scheme(SPG/2021/003801) of the Department of Science and Engineering, Govt. of India, for providing the funding for carrying out this research

\bibliographystyle{ACM-Reference-Format}
\bibliography{sample-base}

\appendix

\end{document}